\documentclass{article} 
\usepackage{iclr2026_conference,times}

\usepackage{amsmath,amsfonts,bm}

\def\eqref#1{equation~\ref{#1}}

\def\1{\bm{1}}

\DeclareMathAlphabet{\mathsfit}{\encodingdefault}{\sfdefault}{m}{sl}
\SetMathAlphabet{\mathsfit}{bold}{\encodingdefault}{\sfdefault}{bx}{n}

\usepackage{url}
\usepackage{colortbl}
\usepackage{hyperref}
\usepackage[OT1]{fontenc}
\usepackage{pifont}
\usepackage{graphicx} 
\usepackage{subcaption}
\usepackage{wrapfig}
\usepackage{multirow}
\usepackage{booktabs}
\usepackage{enumitem}
\usepackage{nicematrix}
\usepackage{xcolor}
\usepackage{color}       
\usepackage[misc]{ifsym}
\usepackage{amssymb}
\definecolor{best}{rgb}{1.0, 0.8, 0.6}
\definecolor{best2}{rgb}{1.0, 0.9, 0.85}
\definecolor{skyblue}{rgb}{0.,0.,0.6}
\definecolor{highlight}{rgb}{1.0,1.0,0.7}
\usepackage{arydshln}

\newcommand{\mycooltitle}{DA$^2$: Depth Anything in Any Direction}
\title{\mycooltitle}

\author{
Haodong Li$^{123\S}$, 
Wangguandong Zheng$^{1}$, 
Jing He$^{3}$, 
Yuhao Liu$^{1}$, 
Xin Lin$^{2}$, 
Xin Yang$^{34}$, 
\\
\ 
\textbf{Ying-Cong Chen}$^{34\ddagger}$\textbf{,} 
\textbf{Chunchao Guo}$^{1\ddagger}$
\\
$^{1}$Tencent Hunyuan $^{2}$UC San Diego $^{3}$ HKUST(GZ) $^{4}$ HKUST
\\
{
\fontfamily{cmtt}\selectfont hal211@ucsd.edu; yingcongchen@ust.hk; chunchaoguo@gmail.com
} 
}

\newcommand{\thetawithspace}{\hspace{.1mm}\theta}

\newcommand{\customfootnotetext}[1]{%
  \begingroup
    \renewcommand{\thefootnote}{}
    \footnotetext{#1}%
  \endgroup
}

\iclrfinalcopy 
\begin{document}

\maketitle

\begin{figure}[h]
    \centering
    \vspace{-10mm}
    \includegraphics[width = 0.98\linewidth]{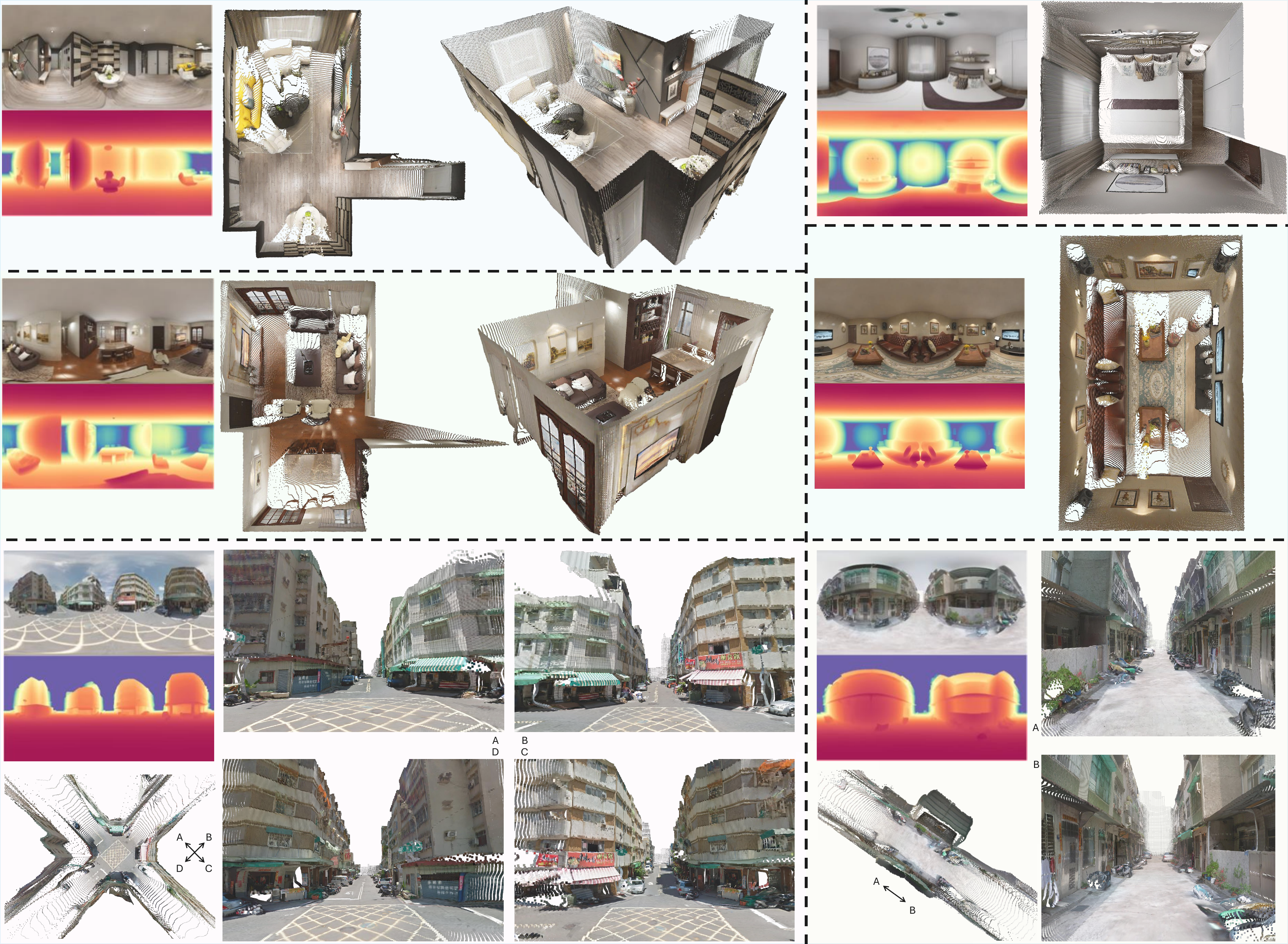}
\caption{Teaser of DA$^2$.~Powered by large-scale training data from our panoramic data curation engine, and the distortion-aware SphereViT, DA$^2$ predicts dense distance from a single 360$^\circ$ panorama, with remarkable geometric fidelity.
The reconstructed 3D structures exhibit sharp geometric details and robust performance across diverse scenes, highlighting DA$^2$'s strong zero-shot generalization.}
\label{fig:teaser}
\end{figure}

\customfootnotetext{\hspace{-1.3mm}{$^\S$Work primarily done during an internship at Tencent Hunyuan. $^{\ddagger}$Corresponding author.}} 

\vspace{-0.1cm}
\begin{abstract}
\vspace{-0.1cm}

Panorama has a full FoV (360$^\circ\times$180$^\circ$), offering a more complete visual description than perspective images.
Thanks to this characteristic, panoramic depth estimation is gaining increasing traction in 3D vision.
However, due to the scarcity of panoramic data, previous methods are often restricted to in-domain settings, leading to poor zero-shot generalization.
Furthermore, due to the spherical distortions inherent in panoramas, many approaches rely on perspective splitting (\textit{e.g.}, cubemaps),
which leads to suboptimal efficiency.
To address these challenges, we propose $\textbf{DA}$$^{\textbf{2}}$: $\textbf{D}$epth $\textbf{A}$nything in $\textbf{A}$ny $\textbf{D}$irection, an accurate, zero-shot generalizable, and fully end-to-end panoramic depth estimator.
Specifically, for scaling up panoramic data, we introduce a data curation engine for generating high-quality panoramic depth data from perspective, and create $\sim$543K panoramic RGB-depth pairs, bringing the total to $\sim$607K.
To further mitigate the spherical distortions, we present SphereViT, which explicitly leverages spherical coordinates to enforce the spherical geometric consistency in panoramic image features, yielding improved performance.
A comprehensive benchmark on multiple datasets clearly demonstrates DA$^{2}$'s SoTA performance, with an average 38\% improvement on AbsRel over the strongest zero-shot baseline.
Surprisingly, DA$^{2}$ even outperforms prior in-domain methods, highlighting its superior zero-shot generalization.
Moreover, as an end-to-end solution, DA$^{2}$ exhibits much higher efficiency over fusion-based approaches.
Both the code and the curated panoramic data has be released.
Project page: \href{https://depth-any-in-any-dir.github.io/}{\texttt{\textcolor{skyblue}{\underline{depth-any-in-any-dir.github.io}}}}.

\end{abstract}

\section{Introduction}
\label{sec:intro}
\vspace{-1mm}

Unlike the commonly used perspective images, panorama offers an immersive 360$^\circ\times$180$^\circ$ view, capturing visual content from \textit{any direction}.
This wide FoV makes panorama an essential visual representation in computer vision, empowering a variety of exciting applications, such as AR/VR~\citep{panoarvr} and immersive visual generation~\citep{yang2025layerpano3d,kalischek2025cubediffrepurposingdiffusionbasedimage}.
However, immersive visual (2D) experiences alone are not enough.
To push the new frontier of
panoramic application scenarios, 
high-quality depth (3D) information from panoramas is crucially needed for 3D reconstruction and more advanced features such as 3D scene generation~\citep{Matrix3D,li4k4dgen,lugenex}, physical simulation~\citep{shah2025virtual}, etc.
Inspired by this, we focuses on estimating scale-invariant\footnote{Please see \textit{Supp}'s Sec.~\ref{supp:depth_category} for discussions on: metric, scale-invariant (biased), and affine-invariant (relative).} distance\footnote{We acknowledge the distinction between distance ($d=\sqrt{x^2+y^2+z^2}$) and depth ($d=z$). We focus on scale-invariant distance prediction. Please allow us to use ``depth'' occasionally for readability and fluency.} from each panorama pixel to the sphere center (\textit{i.e.}, the 360$^\circ$ camera) in an end-to-end manner, with high-fidelity and strong zero-shot generalization.

\begin{figure}[t]
    \centering
    \includegraphics[width=\linewidth]{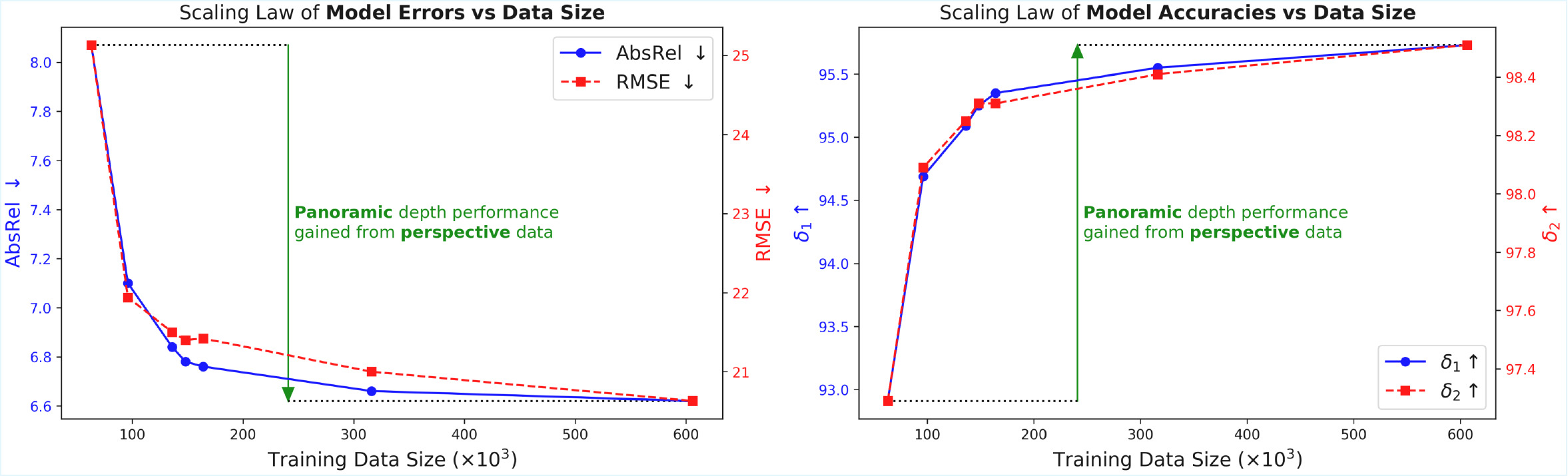}
    \caption{Scaling-law curves of model performance vs data size.
    Native, high-quality panoramic data is scarce, constraining the zero-shot generalization of panoramic depth estimators.
    With our data curation engine, DA$^2$ achieves steadily and clearly higher performance as more perspective depth data are converted to panoramic form.
    Detailed numerical results are provided in Tab.~\ref{tab:scalinglaw}.
    }
    \vspace{-2mm}
    \label{fig:scalinglaw}
\end{figure}

\vspace{-.5mm}
Panoramic depth estimation is particularly valuable for applications requiring comprehensive spatial awareness.
However, capturing or rendering panoramas is much more challenging than perspective images, panoramic depth data is much more limited in both quantity and diversity.
Consequently, early methods were largely trained and tested in in-domain settings, with highly limited zero-shot generalization.
Given the wealth of high-quality perspective depth data, is it possible to transform them into panoramic?
Motivated by this, we propose a data curation engine, transforming perspective samples into high-quality panoramic data.
Concretely, given a perspective RGB image with known horizontal and vertical FoVs, we first apply Perspective-to-Equirectangular (P2E) projection to map the image onto the spherical space.
However, due to the limited FoV of perspective images (with a typical horizontal range of 70$^\circ-$90$^\circ$), only a small portion of the spherical space can be covered (as \colorbox{highlight}{highlighted} in Fig.~\ref{fig:method_data}'s left sphere).
Thus, such a P2E projected image can be viewed as an ``incomplete'' panorama.
Then, panoramic out-painting will be performed to generate a ``complete'' panorama to match the input of our model, using an image-to-panorama out-painter: FLUX-I2P~\citep{flux2024,hunyuanworld2025tencent}.
For the associated GT depth, we apply only the P2E projection \textit{without} out-painting, due to concerns on the \textit{absolute accuracy} of out-painted depth.
Overall, this data curation engine substantially boosts the quantity and diversity of panoramic data, and significantly strengthens the zero-shot performance of DA$^{2}$, as shown in Fig.~\ref{fig:scalinglaw} and Tab.~\ref{tab:scalinglaw}.

\vspace{-.5mm}
Panoramas typically use equirectangular projection (ERP)\footnote{ERP can represent a full vertical FoV (\textit{i.e.}, 180$^\circ$). If smaller than 180$^\circ$, cylindrical projection can be used, such as the panoramic camera mode in mobile phones. Both can present a full horizontal FoV (\textit{i.e.}, 360$^\circ$).} to represent the 360$^\circ\times$180$^\circ$ visual space.
However, a 3D spherical space cannot be ``losslessly'' projected onto a 2D plane.
During the sphere-to-plane projection, distortions and stretching are inevitable, particularly near the poles.
This spherical distortion is analogous to the challenge in world map projection, where you can never accurately express both the areas and shapes of each land. To mitigate the impact of spherical distortion, inspired by the positional embeddings in Vision Transformers (ViTs), we propose SphereViT---the main backbone of DA$^{2}$.
Specifically, from the layout of ERP, we first compute the spherical angles (azimuth and polar) of each pixel in the camera-centric spherical coordinates.
After that, we expand this two-channel angle field into the image feature dimension using sine-cosine basis embedding, forming the Spherical Embedding.
Since all panoramas have the same full FoV,
this spherical embedding can be fixed and reusable.
Therefore, to inject spherical awareness, it's only necessary to let the image feature ``attend'' to the spherical embedding, but not vice versa---the spherical embedding doesn't need to be further refined.
Consequently, rather than adding positional embeddings onto the image features before self-attention, as in standard ViTs~\citep{vaswani2017attention,vit}, SphereViT uses cross-attention: image features are regarded as queries and the spherical embeddings as keys and values.
This design lets the image feature explicitly attend to the panorama's spherical geometry, yielding distortion-aware representations and improved performance.

\vspace{-.5mm}
To validate DA$^{2}$, we conduct a comprehensive benchmark on scale-invariant distance combining multiple well-recognized evaluation datasets.
However, due to the scarcity of panoramic data,
existing zero-shot approaches in panoramic depth estimation are limited, whereas in perspective, there exist many powerful zero-shot methods.
Therefore, to ensure a more fair and comprehensive comparison,
following the \href{https://github.com/microsoft/moge?tab=readme-ov-file#360-panorama-images--moge-infer_panorama}{\textcolor{skyblue}{\underline{panoramic depth estimation pipeline}}} proposed by~\cite{wang2025moge,wang2025moge2}, we also benchmark DA$^{2}$ against prior zero-shot perspective depth estimators~\citep{hu2024metric3d,yin2023metric3d,piccinelli2024unidepth,piccinelli2025unidepthv2,wang2025vggt,wang2025moge,wang2025moge2,bhat2023zoedepth,yang2024depthanything,yang2024depthanythingv2,he2024lotus}, 
The results in Tab.~\ref{tab:main} clearly demonstrate DA$^{2}$'s SoTA performance, with an average 38\% improvement on AbsRel over the strongest zero-shot baseline.
Notably, it even surpasses prior in-domain methods, further underscoring its superior generalization ability.
Beyond that, DA$^{2}$ seamlessly supports various applications, such as panoramic multi-view reconstruction, home decoration, and robotics simulation (please see our \textit{Supp}'s Sec.~\ref{supp:sec:app}). \textbf{Our key contributions are:}

\vspace{-2mm}
\begin{itemize}[itemsep=0.1ex] 
    \item \textbf{Panoramic data curation engine.}~We introduce a data curation engine that generates high-quality panoramic depth data from perspective data, greatly scaling up the panoramic depth training data and substantially improving the zero-shot generalization ability of DA$^2$.
    \item \textbf{SphereViT.}~We propose SphereViT---the primary backbone of DA$^2$. By directly leveraging the spherical coordinates of panoramas, SphereViT effectively mitigates the impact of spherical distortions and enhances the spherical geometry awareness of image features.
    \item \textbf{Comprehensive benchmark.}~Both zero-shot / in-domain, panoramic / perspective methods are compared to build a comprehensive benchmark for panoramic depth estimation.
    \item \textbf{SoTA performance.}~Experimental results clearly demonstrate DA$^{2}$'s SoTA performance. DA$^{2}$ even beats prior in-domain methods. It also enables many downstream applications.
\end{itemize}

\section{Related Works}


\subsection{Perspective Depth Estimation}
\label{subsec:per_de}

Perspective depth estimation is being advanced very rapidly.
Metric and scale-invariant depth models, driven by large-scale training data, have achieved strong results, like
UniDepth~\citep{piccinelli2024unidepth,piccinelli2025unidepthv2}, 
Metric3D~\citep{hu2024metric3d,yin2023metric3d}, DepthPro~\citep{bochkovskiydepth}, and MoGe~\citep{wang2025moge,wang2025moge2}.
Relative depth models also benefit greatly from scaling up the training data, like DepthAnything~\citep{yang2024depthanything,yang2024depthanythingv2}.
Another line of work fine-tunes massively pre-trained generative models, \textit{e.g.}, Stable Diffusion~\citep{rombach2022high,ho2020denoising,he2024disenvisioner,li2024discene,liang2024luciddreamer,gu2024dome}, FLUX~\citep{flux2024,yang2025advancing}, with limited high-quality data, also yielding impressive results~\citep{marigold,he2024lotus,wang2025jasmine,li2024bi}.
Despite these remarkable advances, perspective methods remain constrained by the limited FoV and cannot estimate depth in \textit{all directions} simultaneously.
In contrast, DA$^2$ targets full FoV depth estimation with strong zero-shot generalization.

\begin{figure}[t]
    \centering
    \includegraphics[width=\linewidth]{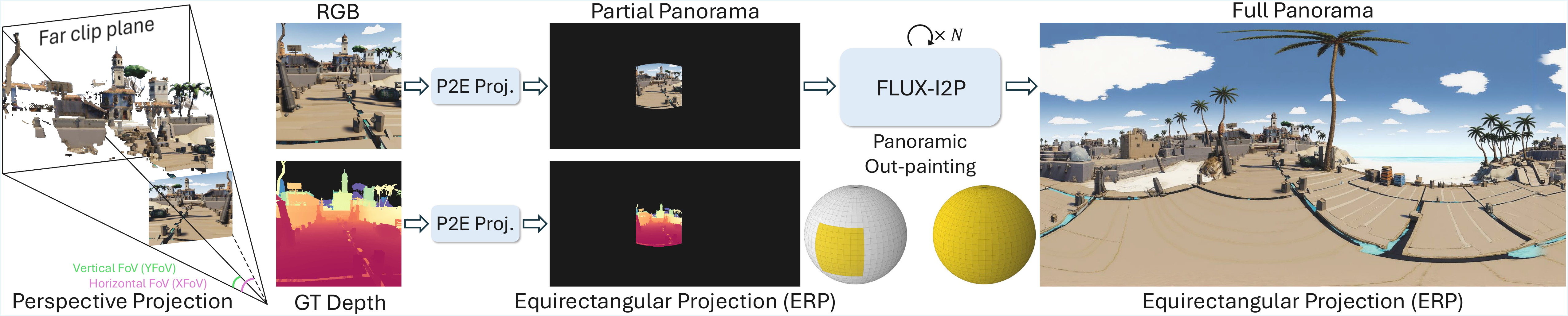}

\vspace{-3mm}
\caption{Panoramic data curation engine.
This module converts large-scale, high-quality perspective RGB–depth pairs into \textit{full} panoramas through P2E projection and panoramic out-painting using FLUX-I2P.
It dramatically scales up the panoramic depth training data, forming a solid training data foundation for DA$^2$.
The \colorbox{highlight}{highlighted} area on the spheres indicate the FoV coverage.}
\vspace{-4mm}
\label{fig:method_data}
\end{figure}

\subsection{Panoramic Depth Estimation}
\label{subsec:pano_de}

\noindent\textbf{In-domain.}~Due to the scarcity of panoramic data, most existing methods are constrained to in-domain settings.
Network designs have evolved from CNNs~\citep{zioulis2018omnidepth,zhuang2022acdnet}) to ViTs~\citep{shen2022panoformer,yun2023egformer}.
Pipeline designs are mainly aimed to mitigate the spherical distortions inherent in panoramas.
Many approaches fuse features from both the ERP (1 panorama) and cubemap (6 perspectives) projections~\citep{wang2020bifuse,jiang2021unifuse,wang2022bifuse++,li2022omnifusion,ai2023hrdfuse,wang2024depthanywhere}.
For alternative solutions, SliceNet~\citep{pintore2021slicenet} and HoHoNet~\citep{sun2021hohonet} use RNNs or LSTMs along longitudes.
SphereDepth~\citep{yan2022spheredepth}, Elite360D~\citep{ai2024elite360d}, HUSH~\citep{lee2025hush} introduce spherical icosahedral meshes and spherical harmonics.
While effective, these strategies still require additional modules, making them less streamlined and efficient.
DA$^2$ introduces SphereViT to handle the spherical distortions in an end-to-end manner, without extra modules.

\noindent\textbf{Zero-shot.}
With the rise of zero-shot perspective depth estimators, there has been a trend toward developing zero-shot depth estimators for panoramas.
360MonoDepth~\citep{rey2022360monodepth} blends tangent perspective depths predicted by MiDaS~\citep{ranftl2020towards} on an icosahedral mesh, but suffers from multi-view inconsistencies.
PanDA~\citep{cao2025panda} leverages \href{https://www.youtube.com/watch?v=JX3VmDgiFnY}{\textcolor{skyblue}{\underline{M$\ddot{\text{o}}$bius transformation}}}-based data augmentation for self-supervision.
UniK3D~\cite{piccinelli2025unik3d} separately predicts camera rays and distance maps, can generalize on various cameras.
But their performance remains suboptimal, due to limited panoramic data: $\sim$20K labeled and $\sim$92K unlabeled in PanDA, $\sim$29K in UniK3D.
DepthAnyCamera~\citep{DepthAnyCamera} projects perspective images with various horizontal FoVs (20$^\circ-$124$^\circ$, $\ll$360$^\circ$) into spherical space, can also generalize on various cameras. But its performance still remains constrained by the incomplete FoVs.
In contrast, DA$^2$ introduces a panoramic data curation engine, significantly boosting the quantity and diversity of panoramic data from available perspective data, yielding a clearly enhanced zero-shot generalization performance.

\vspace{-0.1cm}
\section{Methodology}
\vspace{-0.1cm}

This section presents the methodology of DA$^{2}$ in detail, covering the panoramic data curation engine (Sec.~\ref{subsec:data}) and SphereViT with its training loss functions (Sec.~\ref{subsec:model}).

\vspace{-0.1cm}
\subsection{Panoramic Data Curation Engine}
\label{subsec:data}

\textit{``The quality of your data determines the ceiling of your ambitions.''}~\citep{SurgeAI}

\begin{figure}[t]
    \centering
    \includegraphics[width=\linewidth]{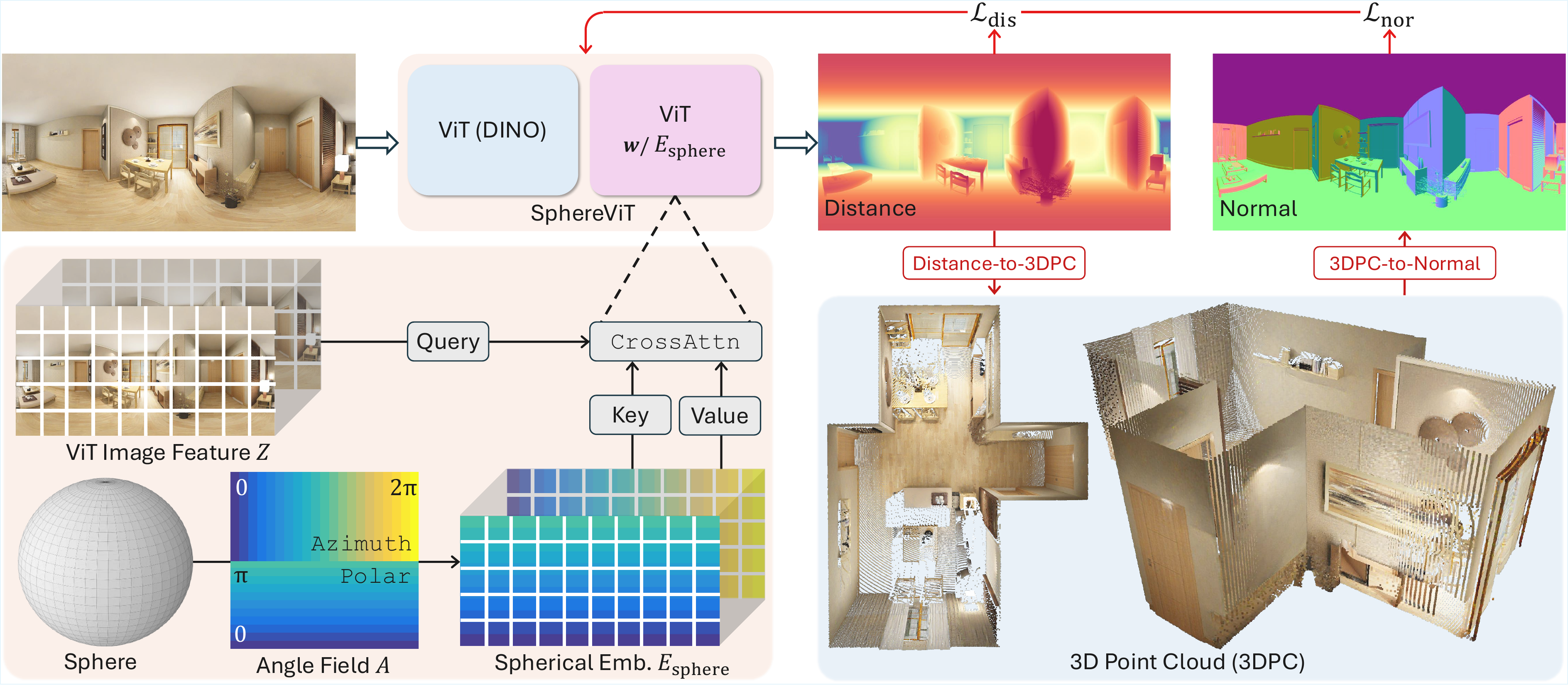}

\caption{The architecture of SphereViT and training losses.
By leveraging the spherical embedding $E_\text{sphere}$, which is explicitly derived from the spherical coordinates of panoramas, SphereViT produces distortion-aware image features, yielding more accurate geometrical estimation for panoramas.
The training supervision combines a distance loss $\mathcal{L}_\text{dis}$ for globally accurate distance values and a normal loss $\mathcal{L}_\text{nor}$ for locally smooth and sharp surfaces.
The effect of $\mathcal{L}_\text{nor}$ is ablated in Fig.~\ref{fig:abl_E_sphere_and_L_nor} (b) and Tab.~\ref{tab:ablation}.
}
\vspace{-2mm}
\label{fig:method_model_loss}
\end{figure}

Due to the scarcity of high-quality panoramic data, existing panoramic depth estimators are often trained and evaluated within specific domains, greatly restricting their zero-shot generalization ability and real-world applicability.
Thus, the very first goal of this work is to scale up the panoramic data and build a strong data foundation for DA$^2$.
Motivated by this, we propose a perspective-to-panoramic data curation engine that generates high-quality panoramic data from perspective data.

As illustrated in Fig.~\ref{fig:method_data}, the inputs of the panoramic data curation engine are a perspective image sized $(W_\text{per},H_\text{per})$ and its FoVs, \textit{i.e.}, XFoV and YFoV.
XFoV represents the coverage of this perspective image in the azimuth field $|\phi_l - \phi_r|$ and YFoV denotes the coverage in the polar angle field $|\theta_u - \theta_d|$.
At first, P2E projection will be performed to map the perspective image onto the spherical space.
Specifically, we start by obtaining the focal lengths from both FoVs: 
\begin{equation}
f_x=\frac{W_\text{per}}{2 \times \tan \left(\frac{\mathrm{FoV}_x}{2}\right)},\quad f_y=\frac{H_\text{per}}{2 \times \tan \left(\frac{\mathrm{FoV}_y}{2}\right)}.
\end{equation}
Then, the 3D vector $\mathbf{d}$ and its unit vector $\hat{\mathbf{d}}$ from the perspective camera to each 2D pixel $(x,y)$ of the perspective image ($x\in [0,W_\text{per}-1],y\in[0,H_\text{per}-1]$) are given by:
\begin{equation}
\mathbf{d}=[\frac{(x-\frac{W_\text{per}-1}{2})}{f_x}, \frac{(y-\frac{H_\text{per}-1}{2})}{f_y}, 1],\quad\hat{\mathbf{d}}=\frac{\mathbf{d}}{|\mathbf{d}|}.
\end{equation}
Then, in the \textit{spherical space}, the azimuth $\phi$ (longitude) and polar $\theta$ (colatitude) angles of $\hat{\mathbf{d}}$ are:
\begin{equation}
\phi=\operatorname{atan2}(\hat{\mathbf{d}}_x, \hat{\mathbf{d}}_z)+\phi_c,\quad \theta=\arccos (\hat{\mathbf{d}}_y)+\theta_c,
\end{equation}
where $(\phi_c, \theta_c)$ denote the spherical coordinates of the perspective image's optical center, used as offsets to obtain the absolute longitude and colatitude of each pixel.
After that, the mapped pixel position $(u, v)$ on the ERP image (\textit{i.e.}, panorama) sized $(W_\text{pano},H_\text{pano})$ is given by:
\begin{equation}
u=\frac{\phi}{2\pi} W_\text{pano},\quad v=\frac{\theta}{\pi} H_\text{pano},
\end{equation}
where $\phi\in[0,2\pi),\theta\in[0,\pi]$.
After P2E projection, due to the limited FoV of perspective images, only a small portion of the sphere can be covered, as \colorbox{highlight}{highlighted} in Fig.~\ref{fig:method_data}'s left sphere.
This incompleteness leads to suboptimal performance:
1) the model lacks global context since it never observes the full views of panoramic images, particularly near the poles; and
2) spherical distortions vary significantly between the equator and poles, with severe stretching occurring at high latitudes.

Thus, following~\citep{hunyuanworld2025tencent}, 
the second step of our data curation engine adopts a LoRA~\citep{hu2022lora} fine-tuned FLUX model named FLUX-I2P for panoramic out-painting, generating ``full'' panoramas from the ``partial'' panoramas.
Earlier panoramic out-painting methods~\citep{gao2024opa,feng2023diffusion360} often exhibited spatial inconsistencies, especially near the poles and the left–right seam.
To address this, FLUX-I2P concatenates image features with the spherical coordinates (azimuth $\phi$ and polar $\theta$) along the channel dimension before feeding them into the Diffusion Transformer (DiT)~\citep{Peebles2022DiT}, to improve the spatial coherence.
For the GT depth associated with the perspective image, we apply only the P2E projection \textit{without} panoramic out-painting, because the \textit{absolute accuracy} of out-painted depth is hard to guarantee.
As ablated in Tab.~\ref{tab:ablation}, although the panoramic out-painting on the P2E projected GT depth is not performed, FLUX-I2P's panoramic out-painting on the RGB images clearly improves the panoramic depth estimation performance by a large margin, demonstrating its significance in our panoramic data curation engine.

\subsection{SphereViT \& Training Losses}
\label{subsec:model}

This data curation engine creates $\sim$543K panoramic samples, scales the total from $\sim$63K to $\sim$607K ($\sim$10 times), significantly addressing the data scarcity issue that causes poor generalization.
Here we focus on DA$^2$'s model structure and training, to effectively learn from the greatly scaled-up data.


Recently, ViT-based depth models have achieved great success~\citep{wang2025moge,wang2025moge2,yang2024depthanything,yang2024depthanythingv2,piccinelli2025unik3d}, where positional embeddings (PE) are crucial for encoding spatial information.
For perspectives, PE is typically derived from the 2D $(x, y)$ pixel coordinates.
However, for panoramas, pixel coordinates $(u, v)$ correspond to spherical coordinates (longitude $\phi$ and latitude $\theta$).
The spherical nature introduces non-uniformity: high-latitude regions (near the poles) are stretched, while low-latitude regions (near the equator) are compressed.
Conventional 2D PE cannot account for this spherical distortion, limiting the model's spherical spatial understanding.
To address this, many approaches fuse features from both the ERP (1 panorama) and cubemap (6 perspectives) projections or employ auxiliary modules, introducing inefficiencies and complexity.
In contrast, DA$^2$ aims to handle the distortions more simply and efficiently, without extra modules.

To this end, DA$^2$ proposes SphereViT, as illustrated in Fig.~\ref{fig:method_model_loss}. SphereViT leverages the spherical coordinates of panoramas to efficiently and explicitly inject spherical-awareness into the ViT image features, yielding distortion-aware representations and improved performance.
Specifically, we first compute the azimuth and polar angles $(\phi,\theta)$ of each pixel $(u, v)$ in an ERP image sized $(W,H)$:
\begin{equation}
    \phi = 2\pi\times\frac{u}{W},\quad \theta = \pi\times\frac{v}{H}.
    \label{eq:af}
\end{equation}
Then, given the image feature $Z\in\mathbb{R}^{(H^\prime\times W^\prime)\times D}$, where $W^\prime=\frac{W}{P}, H^\prime=\frac{H}{P}$ and $P$ is patch size, we resize and flatten this two-channel angle field $A\in\mathbb{R}^{H\times W\times 2}$ (Eq.~\ref{eq:af}) into $A^\prime\in\mathbb{R}^{(H^\prime\times W^\prime)\times 2}$.
Motivated by the PE mechanism of ViT, sine-cosine embedding is utilized to expand $A^\prime$'s channel from $2$ to the image feature dimension $D$.
Concretely, we first define a series of coefficients $\{2^{d_n}\}_{n=1}^{D^\prime}$, where $D^\prime= \frac{D}{4},d_n = \frac{(n-1)\log_2(H^\prime)}{D^\prime}$.
Then, for each two-channel unit of $A^\prime$: $A^\prime_{i,j}=[\phi_i, \theta_j]\in\mathbb{R}^{\text{1}\times\text{2}}$, where $i\in[0,W^\prime-1],j\in[0,H^\prime-1]$, we transpose and multiple it with the coefficients:
\begin{equation}
\begin{bmatrix}
    \phi_i\\
    \theta_j
\end{bmatrix}
\times
\begin{bmatrix}
    2^{d_1}&2^{d_2}&\cdots&2^{d_{D^\prime}}
\end{bmatrix}
=
\begin{bmatrix}
2^{d_1}\phi_i&2^{d_2}\phi_i&\cdots&2^{d_{D^\prime}}\phi_i\\
2^{d_1}\theta_j&2^{d_2}\theta_j&\cdots&2^{d_{D^\prime}}\theta_j
\end{bmatrix}.
\label{eq:before_before_E}
\end{equation}
Eq.~\ref{eq:before_before_E}'s result is shaped $2\times D^\prime$.
We then apply the sine-cosine embedding on each unit of this matrix:
\begin{equation}
\setlength{\arraycolsep}{4.6pt}
\begin{bmatrix}
[\sin(2^{d_1}\phi_i), \cos(2^{d_1}\phi_i)]^{\top}&
[\sin(2^{d_2}\phi_i), \cos(2^{d_2}\phi_i)]^\top&\cdots&
[\sin(2^{d_{D^\prime}}\phi_i),\cos(2^{d_{D^\prime}}\phi_i)]^\top\\
[\sin(2^{d_1}\thetawithspace_j),\cos(2^{d_1}\thetawithspace_j)]^\top&
[\sin(2^{d_2}\thetawithspace_j),\cos(2^{d_2}\thetawithspace_j)]^\top&\cdots&
[\sin(2^{d_{D^\prime}}\thetawithspace_j),\cos(2^{d_{D^\prime}}\thetawithspace_j)]^\top
\end{bmatrix}.
\label{eq:before_E}
\end{equation}
Eq.~\ref{eq:before_E} has a shape of $2\times D^\prime\times 2$.
Now, the flattened transformation of Eq.~\ref{eq:before_E}---with a dimension of $2\times D^\prime\times 2=D$---is the unit $(i,j)$ of the Spherical Embedding $E_\text{sphere}\in\mathbb{R}^{(H^\prime\times W^\prime)\times D}$.

As discussed in Sec.~\ref{sec:intro}, all panoramas share the same 360$^\circ\times $180$^\circ$ FoV, so the spherical embedding is fixed, reusable, and doesn't need to be further refined.
Thus, to inject spherical awareness, it's only necessary to let image features $Z$ ``attend'' to the embedding $E_\text{sphere}$, but not vice versa.
Accordingly, SphereViT replaces the usual self-attention (after addition: $Z + E_\text{sphere}$) with cross-attention, where image features $Z$ serve as queries and the spherical embeddings $E_\text{sphere}$ act as keys and values:
\begin{equation}
\texttt{CrossAttn}\left(Z, E_{\text {sphere }}\right)=\operatorname{SoftMax}\left(\frac{Z W_Q\left(E_{\text {sphere }} W_K\right)^{\top}}{\sqrt{D_k}}\right)\left(E_{\text {sphere }} W_V\right),
\end{equation}
where $W_Q, W_K, W_V \in \mathbb{R}^{D \times D_k}$ are learnable projection matrixs, and $Z,E_\text{sphere}\in\mathbb{R}^{(H^\prime\times W^\prime)\times D}$.
This cross-attention with spherical embedding $E_\text{sphere}$ allows the image features $Z$ to ``learn'' the underlying spherical structures of the panoramas, producing distortion-aware representations and leading to clearly enhanced geometrical fidelity
as demonstrated in Fig.~\ref{fig:abl_E_sphere_and_L_nor} (a) and Tab.~\ref{tab:ablation}.

\noindent\textbf{Training Losses.}~
DA$^2$'s SphereViT is trained end-to-end to estimate dense, scale-invariant distance $\hat{D}\in\mathbb{R}^{H\times W}$ from a panoramic RGB input $I\in\mathbb{R}^{H\times W\times 3}$.
The supervision combines two terms: a distance loss $\mathcal{L}_{\text{dis}}$ that enforces globally accurate distance values, and a normal loss $\mathcal{L}_{\text{nor}}$ that promotes locally smooth, sharp geometrical surfaces, especially in regions where distance values are similar but surface normals vary significantly.
Concretely, let $\hat{D}$ and $D^\star$ be the predicted and GT distances.
Then the surface normals can be obtained with a distance-to-normal operator \texttt{D2N}, giving $\hat{N}=\texttt{D2N}(\hat{D})$ and $N^\star=\texttt{D2N}(D^\star)$ when GT normals are not directly available.
Since we focuses on scale-invariant distance, $\hat{D}$ is median-aligned before loss computing: $\hat{D}^\text{med}=\hat{D}\times\frac{\operatorname{Median}(D^\star)}{\operatorname{Median}(\hat{D})}$.
While training the SphereViT, we minimize the per-pixel L1 difference for both $\mathcal{L}_{\text{dis}}$ and $\mathcal{L}_{\text{nor}}$:
\begin{equation}
\mathcal{L}_\text{dis} = \frac{1}{|\Omega|}
\sum_{p\in\Omega}
\bigl|\,\hat{D}^\text{med}_p-D^\star_p\,\bigr|,\quad
\mathcal{L}_{\text{nor}}=
\frac{1}{|\Omega|}
\sum_{p\in\Omega}
\bigl|
\,\hat{N}_p-N^\star_p\,
\bigr|,
\end{equation}
where $\Omega$ is the set of valid pixels.
For $\mathcal{L}_\text{nor}$, we prefer the L1 norm over the commonly-used angular discrepancy $1-\langle \hat{N}_p, N_p\rangle$ as the latter may introduce gradient collapse and destabilize training.
The total loss is a weighted sum: $\mathcal{L}=\lambda_\text{d}\mathcal{L}_\text{dis} + \lambda_\text{n}\mathcal{L}_\text{nor}$, where $\lambda_{\text{d}}$ and $\lambda_{\text{n}}$ are scalar weights.

\section{Experiments}

\setlength{\tabcolsep}{0.5pt}
\begin{table}[t]
\centering
\fontsize{7.2pt}{8.2pt}\selectfont
\caption{Quantitative comparison. For a fair and comprehensive benchmark, we include both zero-shot / in-domain, panoramic / perspective approaches.
The \colorbox{best}{best} and \colorbox{best2}{second best} performances are highlighted (in \textit{zero-shot} setting).
In \textit{all} settings (both zero-shot and in-domain), the \textbf{best} and \underline{second best} performances are bolded and underlined.
DA$^2$ outperforms all other methods no matter in zero-shot or all settings, particularly showing large gains under the zero-shot setting.
Median alignment (scale-invariant) is adopted by default.
$^{\vartriangle}$: Affine-invariant alignment (scale and shift-invariant), for prior \textit{relative} depth estimators: DepthAnything v1v2~\citep{yang2024depthanything,yang2024depthanythingv2}, Lotus~\citep{he2024lotus}, and PanDA~\citep{cao2025panda}. We also report PanDA's results in median alignment for fairness.
$^\star$: Implemented by ourselves (code will be released).
The unit is percentage (\%).}
\vspace{-2mm}
\begin{NiceTabular}{ccccccccccccccccc}
\toprule
\multirow{2}{4.2em}{Categories} & \multirow{2}{3.1em}{Method} & 
\multicolumn{4}{c}{Stanford2D3D} & 
\multicolumn{4}{c}{Matterport3D} & 
\multicolumn{4}{c}{PanoSUNCG} & 
Rank$\downarrow$ & 
Rank$\downarrow$
\\
\cmidrule(lr){3-6}
\cmidrule(lr){7-10}
\cmidrule(lr){11-14}
& &
AbsRel$\downarrow$ & RMSE$\downarrow$ & $\delta_1\uparrow$ & $\delta_2\uparrow$& 
AbsRel$\downarrow$ & RMSE$\downarrow$ & $\delta_1\uparrow$ & $\delta_2\uparrow$& 
AbsRel$\downarrow$ & RMSE$\downarrow$ & $\delta_1\uparrow$ & $\delta_2\uparrow$& 
Zero-shot&
All\ \ &
\\
\cmidrule(lr){1-1}
\cmidrule(lr){2-2}
\cmidrule(lr){3-6}
\cmidrule(lr){7-10}
\cmidrule(lr){11-14}
\cmidrule(lr){15-15}
\cmidrule(lr){16-16}
\multirow{17}{4.2em}{In-domain}
&OmniDepth & 19.96& 61.52& 68.77& 88.91 & 29.01& 76.43& 68.30& 87.94 & 11.43& 37.10& 87.05& 93.65 &-- & 26.33\\
&FCRN & 18.37& 57.74& 72.30& 92.07 & 24.09& 67.04& 77.03& 91.74 & 9.79& 39.73& 92.23& 96.59  & --& 24.00\\
&BiFuse & 12.09& 41.42& 86.60& 95.80 & 20.48& 62.59& 84.52& 93.19 & 5.92& 25.96& 95.90& 98.23& --& 16.83\\
&EGFormer & 15.28& 49.74& 81.85& 93.38 & 14.73& 60.25& 81.58& 93.90 & --& --& --& --&-- &15.50 \\
&SliceNet & 12.49& 43.70& 83.77& 94.14 & 17.64& 61.33& 87.16& 94.83 & --& --& --& --&-- & 14.17\\
&SphereDepth& 11.58& 45.12& 86.66& 96.42 & 12.05& 59.22& 86.20& 95.19 & --& --& --& --&-- & 12.42\\
&BiFuse++ & 11.17& 37.20& 87.83& 96.49  & 14.24& 51.90& 87.90&  95.17 & \textbf{5.24}& 24.77& \textbf{96.30}& 98.35&-- & 12.17\\
&UniFuse & 11.14& 36.91& 87.11& 96.64 & 10.63& 49.41& 88.97& 96.23 & \underline{5.28}& 27.04& 95.91& 98.25&-- & 11.25\\
&HoHoNet & 10.14& 38.34& 90.54& 96.93 & 14.88& 51.38& 87.86& 95.19 & --& --& --& --& --& 10.33\\
&Elite360D & 11.82& 37.56& 88.72& 96.84 & 11.15& 48.75& 88.15& 96.46 & --& --& --& --&-- & 10.00\\
&PanoFormer& 11.31& 35.57& 88.08& 96.23 & 9.04& 44.70& 88.16& 96.61  & 5.34& \textbf{18.90}& 94.87& \textbf{98.83}  &-- & 9.50\\
&HRDFuse& 9.35& 31.06& 91.40& 97.98 & 9.67& 44.33& 91.62& 96.69 & 6.90& 27.44& 92.15& 97.42 &-- & 9.50\\
&SphereFusion& 8.99& 31.94& 92.57& 97.55  & 11.45& 48.85& 87.01& 96.13 & --& --& --& --&-- &7.92 \\
&ACDNet & 9.84& 34.10& 88.72& 97.04  & 10.10& 46.29& 90.00& 96.78 & --& --& --& --&-- & 7.08\\
&DepthAnywhere & 11.80& 35.10& 91.00& 97.10 & 8.50& --& 91.70& 97.60     &-- &-- &-- &-- &-- & 5.33\\
&OmniFusion& 9.50& 34.74& 89.88& 97.69 & 9.00& 42.61& 91.89& 97.97 & --& --& --& --&-- & 5.00\\
&\underline{HUSH} & \underline{7.82}& 33.32& \underline{93.84}& \textbf{98.49}& \underline{8.38}& 41.64& 92.87& 96.98&-- &-- &-- &-- &-- &\underline{3.67} \\
\cmidrule(lr){1-1}
\cmidrule(lr){2-2}
\cmidrule(lr){3-6}
\cmidrule(lr){7-10}
\cmidrule(lr){11-14}
\cmidrule(lr){15-15}
\cmidrule(lr){16-16}
\multirow{13}{4.2em}{Zero-shot \textcolor{white}{(}(fusion)}
&Lotus-D$^{\star\vartriangle}$ &45.88 & 48.86& 37.67& 68.39& 32.39& 85.86& 48.15& 78.23& 37.96& 77.02& 46.08& 77.41& 17.00& 30.33\\
&Lotus-G$^{\star\vartriangle}$ & 45.08& 47.90& 38.38& 69.18& 31.82& 84.51& 49.11& 78.92& 38.02& 76.82& 46.16& 77.51& 16.17& 29.50\\
&DepthAnything$^{\star\vartriangle}$ & 37.21& 43.41& 47.08& 76.93& 24.46& 66.12& 60.54& 88.32& 24.58& 52.22& 64.86& 90.39& 14.58& 27.42\\
&DepthAnythingv2$^{\star\vartriangle}$ & 36.79& 43.39& 47.66& 76.96& 25.85& 70.67& 58.42& 86.19& 23.90& 50.74& 66.86& 90.89& 14.25& 27.25\\
&ZoeDepth$^\star$ & 17.60& 33.74& 74.26& 92.86& 18.43& 53.46& 72.18& 93.12& 21.16& 44.81& 69.34& 94.45& 11.75&22.58 \\
&360MonoDepth & 16.50& 28.23& 74.56& 92.98& 20.83& 79.09& 65.58& 88.95& 11.43& 28.29& 90.75& 98.12& 10.83& 21.67\\
&VGGT$^\star$ & 18.70& 33.50& 74.08& 83.90& 10.78& 38.80& 88.70& 97.72& 8.43& 25.67& 94.04& 98.19& 8.42& 15.08\\
&Metric3D$^\star$ & 12.93& 20.80& 84.77& 96.52& 14.11& 45.11& 83.09& 96.59& 11.42& 26.95& 90.45& 97.33& 7.67& 15.17\\
&UniDepth$^{\star}$ & 15.06& 20.48& 76.99& 90.34& 11.12& 36.20& 88.66& 97.94& 10.40& 27.29& 92.59& 98.00& 7.50& 13.92\\
&MoGe & 15.81& 25.76& 79.02& 83.32& 10.04& 35.91& 90.80& 98.45& 8.60& 25.80& 93.85& 98.31& 6.33& 12.08\\
&UniDepthv2$^{\star}$ & 13.08& 20.46& 82.12& 89.21& 10.86& 37.68& 88.76& 97.86& 9.74& 25.94& 93.06& 98.30& 6.25& 12.17\\
&Metric3Dv2$^{\star}$ & 11.59& 21.78& 86.07& \cellcolor{best2}97.36& 17.78& 62.55& 72.35& 93.22& 7.30& 24.54& 94.25& 98.25& 6.08& 14.08\\
&MoGev2 & 14.69& 24.24& 79.98& 84.39& 10.34& 36.91& 89.48& 98.24& 8.26& 24.67& 94.15&\cellcolor{best2} 98.52& 5.58& 11.25\\
\cmidrule(lr){1-1}
\cmidrule(lr){2-2}
\cmidrule(lr){3-6}
\cmidrule(lr){7-10}
\cmidrule(lr){11-14}
\cmidrule(lr){15-15}
\cmidrule(lr){16-16}
\multirow{5}{4.2em}{Zero-shot (end2end)}
&PanDA& 48.44 & 53.06 & 33.92 & 51.33 & 37.10& 101.5& 42.51& 67.29& 34.73& 79.69& 44.49& 71.45& 17.50& 30.83\\
&DepthAnyCamera& 15.26& 22.80& 75.47& 92.90& 15.60& 61.85& 77.27& 95.62& 12.78& 27.88& 89.67& 97.85& 9.75& 19.42\\
&PanDA$^{\vartriangle}$& 16.48& 23.64& 73.26& 85.42& \cellcolor{best2}8.88& 33.25& 92.09& 98.26& \cellcolor{best2}6.71& \cellcolor{best2}21.85& \cellcolor{best2}95.42& 98.25&5.33 &10.33 \\
&\cellcolor{best2} UniK3D& \cellcolor{best2}11.31& \cellcolor{best2}\underline{19.72}& \cellcolor{best2}88.94& 95.33&
9.66& \cellcolor{best2}\underline{32.66}& \cellcolor{best2}\underline{93.00}&\cellcolor{best2} \underline{98.58}& 
11.46& 25.38& 90.18& 98.02&\cellcolor{best2} 4.58& 8.75\\
&\cellcolor{best}\textbf{DA}$^{\textbf{2}}$ \textbf{(Ours)}&
\cellcolor{best} \textbf{7.23}&\cellcolor{best} \textbf{14.00}&\cellcolor{best} \textbf{95.45}&\cellcolor{best} \underline{98.38}& 
\cellcolor{best}\textbf{6.67}& \cellcolor{best}\textbf{28.82}& \cellcolor{best}\textbf{95.61}&\cellcolor{best} \textbf{98.60}& 
\cellcolor{best}5.96& \cellcolor{best}\underline{19.07}& \cellcolor{best}\underline{96.12}& \cellcolor{best}\underline{98.55}&
\cellcolor{best}1.00&\textbf{1.67}\\ 
\bottomrule
\end{NiceTabular}
\label{tab:main}
\vspace{-4mm}
\end{table}




\subsection{Experimental Settings}

\label{sec:exp_set}
\noindent\textbf{Training Datasets.}~DA$^2$ is trained using 7 high-quality datasets. 6 perspective: Hypersim~\citep{Hypersim},
Virtual-KITTI 2~\citep{vkitti},
MVS-Synth~\citep{MVSSynth},
UnrealStereo4K~\citep{unrealstereo4k},
3D-Ken-Burns~\citep{3dkenburns},
Dynamic Replica~\citep{DynamicReplica}, totaling 543,425 samples; 1 panoramic: Structured3D~\citep{zheng2020structured3d} (63,097 samples).

\noindent\textbf{Evaluation Datasets \& Metrics.}~For a fair and reproducible comparison, DA$^{2}$ is evaluated on three widely-used, well-recognized benchmarks in panoramic depth estimation: Stanford2D3D-S~\citep{stanford2d3ds} (all splits), Matterport3D~\citep{chang2017matterport3d} (test split), and PanoSUNCG~\citep{PanoSUNCG} (test split), using 2 error metrics (AbsRel, RMSE), and 2 accuracy metrics ($\delta_1$, $\delta_2$).
Please see the implementation details (Sec.~\ref{supp:sec:impl}) and metric formulations (Sec.~\ref{supp:sec:metrics}) in our \textit{Supp}.

\begin{wrapfigure}{rt}{0.605\textwidth}
    \centering
\includegraphics[width=\linewidth]{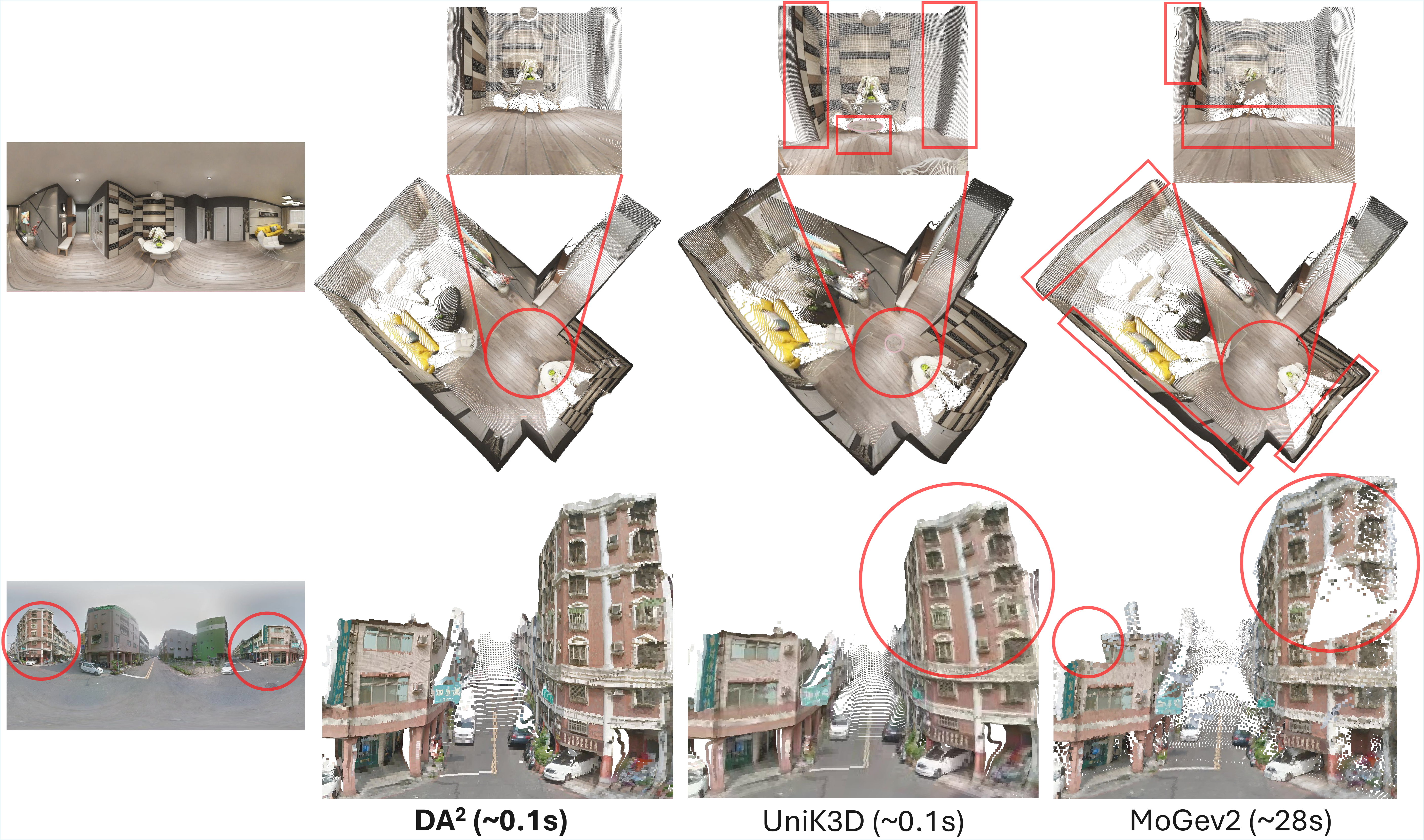}
\vspace{-4mm}
\caption{Qualitative comparisons. Compared with UniK3D and MoGev2, DA$^2$ delivers more accurate geometric predictions and, as an end-to-end approach, achieves significantly higher inference efficiency than fusion-based methods.
}
\label{fig:qc}
\vspace{-5mm}
\end{wrapfigure}

\vspace{-6mm}
\subsection{Quantitative \& Qualitative Comparisons}

Tab.~\ref{tab:main} presents a comprehensive comparison of DA$^2$ with previous SoTA approaches.
Following~\citep{wang2025moge,wang2025moge2}, we also include prior perspective methods for a more thorough comparison.
As demonstrated in Tab.~\ref{tab:main}, DA$^2$ consistently outperforms all other methods across various settings.
Particularly in the zero-shot setting, DA$^2$ shows significant gains over the second-best method by an average of 38\% in AbsRel and 22\% in RMSE, achieving a remarkable average $\delta_1$ of 95.73\% and $\delta_2$ of 98.51\%.
Notably, even as a zero-shot model, DA$^2$ surpasses earlier in-domain methods as well, further underscoring its superior zero-shot generalization ability.

In addition, for better access the DA$^2$'s performance, we also conduct qualitative comparisons with UniK3D~\citep{piccinelli2025unik3d}---the strongest prior zero-shot, end-to-end method, and MoGev2~\citep{wang2025moge2}---the strongest prior zero-shot, fusion-based method, as highlighted in Fig.~\ref{fig:qc}.
Thanks to our data curation engine, DA$^2$ is trained with about 21$\times$ more panoramic data than UniK3D, exhibiting clearly more accurate geometrical predictions.
DA$^2$ continuously yields better results over MoGev2, as its panoramic performance is restricted by the multi-view inconsistencies during fusion, \textit{e.g.}, irregular walls, fragmented buildings, etc.
We also report the inference times: as an end-to-end method, DA$^2$ achieves significantly higher efficiency than fusion-based approaches.

\subsection{Ablation Studies}
\setlength{\tabcolsep}{2.5pt}
\begin{table}[t]
\centering
\footnotesize

\vspace{-3mm}
\caption{Ablation study on training data scaling.
The results show clear, steady performance gains as the size of training data grows.
$^\text{Pano}$ indicates a perspective dataset converted into panoramic through our data curation engine.
The average results across multiple datasets are reported (also in Tab.~\ref{tab:ablation}).
Please see \textit{Supp}'s Sec.~\ref{supp:sec:data} for more discussions about the date curation engine and the curated data.
}
\vspace{-3mm}

\begin{NiceTabular}{cccccccccccc}
\toprule
S3D&HPS$^\text{Pano}$ &VK$^\text{Pano}$ & MVS$^\text{Pano}$& US4K$^\text{Pano}$ &3DKB$^\text{Pano}$ &DR$^\text{Pano}$ & Data Size & AbsRel$\downarrow$ & RMSE$\downarrow$ & $\delta_1\uparrow$ & $\delta_2\uparrow$\\
\cmidrule(lr){1-7}
\cmidrule(lr){8-8}
\cmidrule(lr){9-12}
\checkmark& \ding{55}& \ding{55}& \ding{55}& \ding{55}& \ding{55}& \ding{55}& 63,097& 8.07& 25.13&92.91 &97.29 \\
\checkmark& \checkmark& \ding{55}& \ding{55}& \ding{55}& \ding{55}& \ding{55}& 96,677& 7.10& 21.94& 94.69 &98.09 \\
\checkmark& \checkmark& \checkmark& \ding{55}&\ding{55} &\ding{55} &\ding{55} &136,326 &6.84 &21.50 &95.09 &98.25 \\
\checkmark& \checkmark& \checkmark& \checkmark& \ding{55}&\ding{55} &\ding{55} &148,326 & 6.78&21.40 &95.25 &98.31 \\
\checkmark& \checkmark& \checkmark& \checkmark& \checkmark&\ding{55} &\ding{55} &164,726 &6.76 & 21.42&95.35 &98.31 \\
\checkmark& \checkmark& \checkmark& \checkmark& \checkmark& \checkmark& \ding{55}&316,722 & \underline{6.66}& \underline{21.00}& \underline{95.55}&\underline{98.41} \\
\cmidrule(lr){1-7}
\cmidrule(lr){8-8}
\cmidrule(lr){9-12}
\checkmark& \checkmark& \checkmark& \checkmark& \checkmark& \checkmark& \checkmark& 606,522 & \textbf{6.62}& \textbf{20.63}& \textbf{95.73}& \textbf{98.51}\\
\bottomrule
\end{NiceTabular}
\label{tab:scalinglaw}
\vspace{-2mm}
\end{table}


\setlength{\tabcolsep}{2.6pt}
\begin{table}[t]
\centering
\footnotesize
\caption{Ablation studies on: 1) the panoramic out-painting in the data curation engine, 2) spherical embedding $E_\text{sphere}$ in the SphereViT, and 3) the auxiliary normal loss $\mathcal{L}_\text{nor}$.
The results below demonstrate that each design plays a vital role in achieving the final remarkable performance of DA$^2$.}
\vspace{-3mm}
\begin{NiceTabular}{cccccccc}
\toprule
Pano. Out-painting&Spherical Emb. $E_\text{sphere}$ &Normal Loss $\mathcal{L}_\text{nor}$ & Data Size & AbsRel$\downarrow$ & RMSE$\downarrow$ & $\delta_1\uparrow$ & $\delta_2\uparrow$\\
\cmidrule(lr){1-3}
\cmidrule(lr){4-4}
\cmidrule(lr){5-8}
\ding{55}& \checkmark& \checkmark&606,522 & 7.59 &23.80 &94.12 &97.86 \\
\checkmark&  \ding{55}&\checkmark&606,522 & \underline{6.84} &\underline{20.87} &\underline{95.26} &\underline{98.43} \\
\checkmark& \checkmark&\ding{55}& 606,522 &6.99 &21.53 &95.25 &98.37 \\
\cmidrule(lr){1-3}
\cmidrule(lr){4-4}
\cmidrule(lr){5-8}
\checkmark& \checkmark& \checkmark&606,522 & \textbf{6.62}& \textbf{20.63}& \textbf{95.73}& \textbf{98.51}\\
\bottomrule
\end{NiceTabular}
\label{tab:ablation}
\end{table}

\begin{figure}[h]
\centering\includegraphics[width=0.9\linewidth]{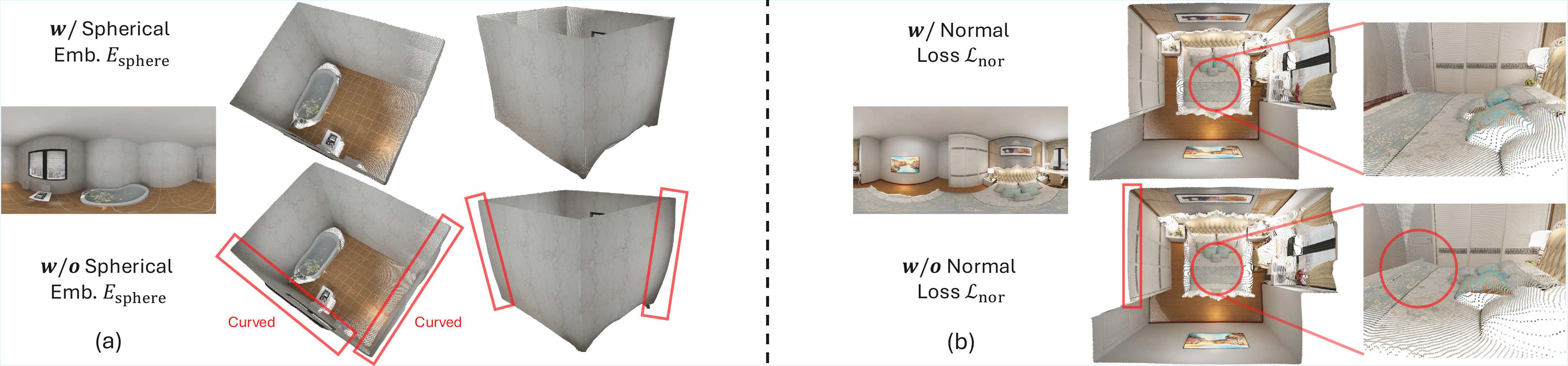}
    \caption{Ablation studies of DA$^2$.
(a) Removing the spherical embedding $E_{\text{sphere}}$ causes curved, distorted geometry.
(b) Omitting the normal loss $\mathcal{L}_{\text{nor}}$ yields rougher surfaces and more artifacts.}
\label{fig:abl_E_sphere_and_L_nor}
\vspace{-7mm}
\end{figure}

\noindent\textbf{Training Data.}~As reported in Tab.~\ref{tab:scalinglaw}, DA$^2$'s performance steadily improves as more perspective depth data converted into panoramic, thanks to our data curation engine.
Fig.~\ref{fig:scalinglaw} further shows rapid gains once the curated perspective data is introduced, with performance gradually converging as the data scales.
Even near convergence, further improvements are still anticipated with additional data.



\noindent\textbf{Panoramic Out-painting.}~It is a crucial step in the panoramic data curation engine, generating \textit{full} RGB panoramas from P2E-projected perspective images (Fig.~\ref{fig:method_data}).
Comparing Tab.~\ref{tab:scalinglaw}'s 1$^\text{st}$ row with Tab.~\ref{tab:ablation}'s 1$^\text{st}$ row, DA$^2$'s performance can be improved only modestly via scaling up the perspective \textbf{\textit{w/o}} panoramic out-painting, yielding a 0.48 gain in AbsRel.
In contrast, incorporating (\textbf{\textit{w/}}) out-painting yields a much larger boost than ``\textbf{\textit{w/o}} out-painting'' ($\sim$3 times), with a 1.45 gain in AbsRel (Tab.~\ref{tab:scalinglaw}'s 1$^\text{st}$ row vs.~Tab.~\ref{tab:ablation}'s last row), clearly showing the importance of panoramic out-painting.

\noindent\textbf{Spherical Embedding.}~We here ablate the impact of spherical embedding $E_\text{sphere}$ in the SphereViT.
As shown in Tab.~\ref{tab:ablation} (2$^\text{nd}$ vs.~last row), including $E_\text{sphere}$ noticeably boosts DA$^2$’s performance.
Fig.~\ref{fig:abl_E_sphere_and_L_nor} (a) further illustrates that incorporating the spherical embedding produces more accurate geometric understandings on panoramas, while its absence often leads to suboptimal performance (\textit{e.g.}, curved walls), highlighting its effectiveness in mitigating the spherical distortions.

\noindent\textbf{Training Losses.}~We further ablate the auxiliary normal loss $\mathcal{L}_\text{nor}$ used for training the SphereViT.
As shown in Tab.~\ref{tab:ablation} (3$^\text{rd}$ vs.~last row), adding $\mathcal{L}_\text{nor}$ boosts DA$^2$'s performance clearly.
Also, as highlighted in Fig.~\ref{fig:abl_E_sphere_and_L_nor} (b), normal supervision yields flatter, smoother, and more coherent geometry, reducing the artifacts that typically appear in ambiguous regions (\textit{e.g.}, corners, edges, and the upper or lower poles), where distance values may be similar but surface normals differ substantially.
\begin{wrapfigure}{tr}{0.5\textwidth}
    \vspace{.5cm}
    \centering
    \includegraphics[width=\linewidth]{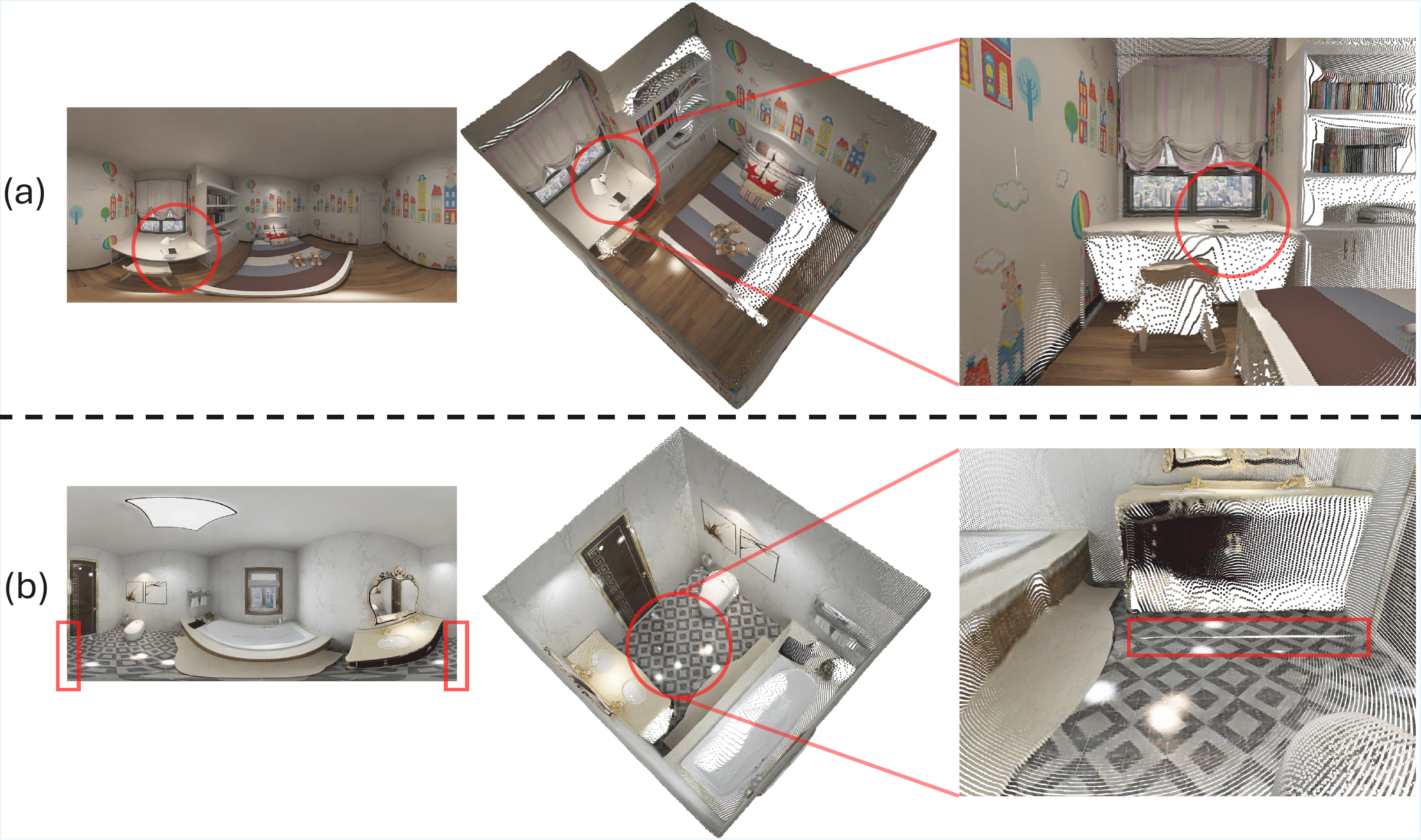}
\caption{DA$^2$'s limitations.
(a) The white lamp's predicted distance is mistakenly aligned with the desk surface.
(b) Visible seams appear along the predictions at lower left–right boundaries.
}
    \label{fig:limitation}
    \vspace{-20mm}
\end{wrapfigure}

\section{Limitation \& Conclusion}

\noindent\textbf{Limitation.}~Despite the strong performance enabled by the large-scale training data thanks to our panoramic data curation engine and distortion-aware SphereViT, DA$^2$ still faces several constraints.
As the training resolution (1024$\times$512) is lower than higher-definition formats such as 2K or 4K, and the curated perspective data provide only partially available GT depth in the spherical space, DA$^2$ may occasionally miss fine details (Fig.~\ref{fig:limitation} (a)) and produce visible seams along the panorama's left–right boundaries (which should ideally be seamlessly aligned), as illustrated in Fig.~\ref{fig:limitation} (b).

\newpage
\noindent\textbf{Conclusion.}~We introduce DA$^2$, an end-to-end, zero-shot generalizable, panoramic distance (scale-invariant) estimator that unites a panoramic data curation engine with the distortion-aware SphereViT.
Trained on over 600K samples ($\sim$543K curated from perspective and $\sim$63K native panoramas), DA$^2$ delivers SoTA zero-shot performance, outperforming prior methods (both zero-shot and in-domain) by a clear margin while remaining efficient and fully end-to-end.
This work shows that scaling up panoramic data and explicitly modeling the spherical geometry enables high-quality and robust 360$^\circ\times$180$^\circ$ geometrical estimation, paving the way for high-fidelity 3D scene applications, \textit{e.g.}, immersive 3D scene creation, AR/VR, robotics simulation, physical simulation, etc.



\section{Acknowledgment}

We sincerely thank Dr.~Hualie Jiang from CUHK(SZ) for providing access to Matterport3D's test split, which is invaluable in enabling the comprehensive benchmark.
We also thank Luigi Piccinelli from ETH Zurich and Ruicheng Wang from USTC and MSRA for their thoughtful discussions.

\section{LLMs in Paper Writing}

LLMs (\textit{e.g.}, GPT-4, GPT-5) were employed solely to polish the grammar and sentence structures, for improving the readability, clarity, and fluency.
They made no contribution to the original research content.
All the scientific and technical content of this paper was written entirely by humans.



\appendix
\newpage

{\LARGE\sc Supplementary Materials of\\\mycooltitle \par}

\vspace{.3cm}
\section{Applications of DA$^2$}
\label{supp:sec:app}

Leveraging its remarkable capability in zero-shot generalizable panoramic depth estimation, DA$^2$ effectively enables a wide range of 3D reconstruction-related applications.


\begin{figure}[!h]
    \centering
    \includegraphics[width=\linewidth]{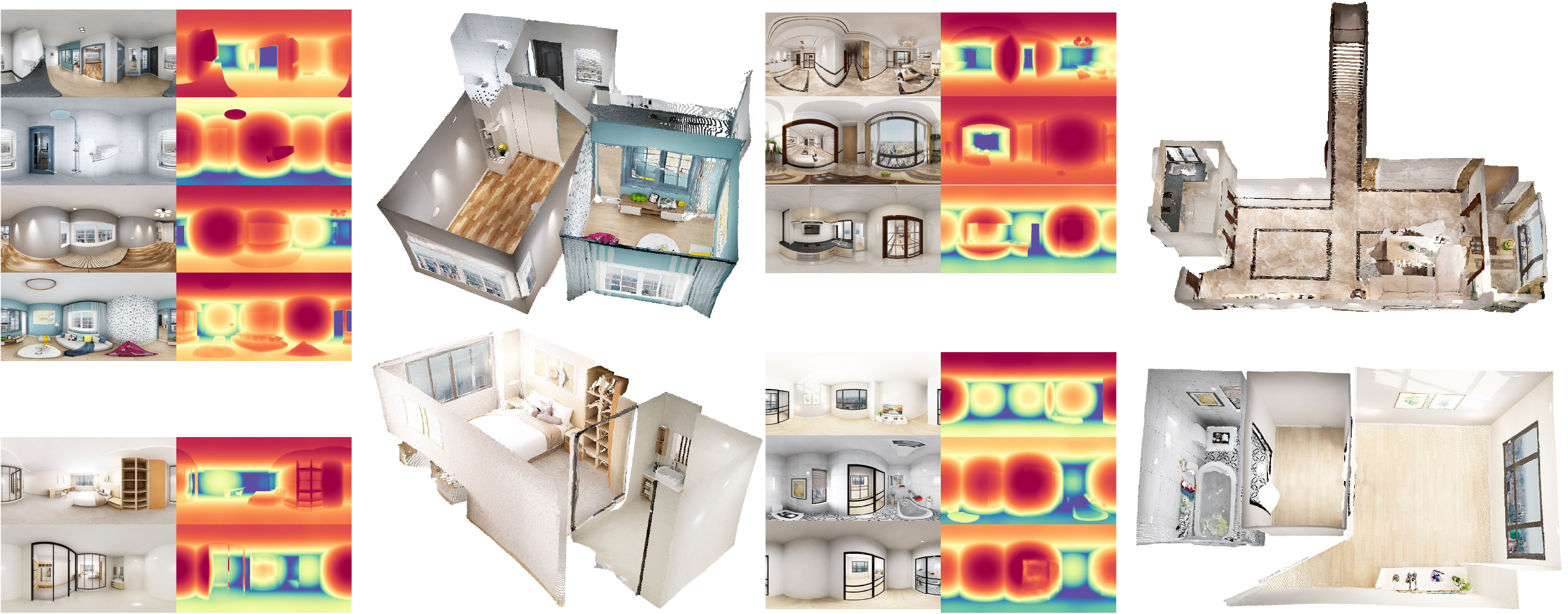}
    \caption{Pano3R: Panoramic Multi-view Reconstruction. Given panoramic images of different rooms from a house / apartment, DA$^2$ enables the reconstruction of a globally aligned 3D point cloud, ensuring the spatial coherence across multiple panoramic views of different rooms.
    }
    \label{fig:application1}
\end{figure}

\subsection{Pano3R: Panoramic Multi-view Reconstruction}
A house / apartment typically consists of multiple distinct rooms, which may exhibit substantial geometric variations. Thanks to the strong zero-shot generalization and high geometric consistency in panoramic depth estimation, DA$^2$ is able to reconstruct a holistic 3D point cloud representation of the indoor layout, leveraging multiple panoramic images captured from different rooms.
As shown in Fig.~\ref{fig:application1}, the rooms can be consistently aligned via simple translation, without requiring any scaling or rotation operations.
This characteristic highlights the robustness and superior geometric consistency of DA$^2$'s depth estimation, enabling seamless alignment of shared structures such as walls and doors, facilitating applications such as VR-based indoor apartment tours and layout visualization.



\begin{figure}[!h]
    \centering
    \includegraphics[width=0.9\linewidth]{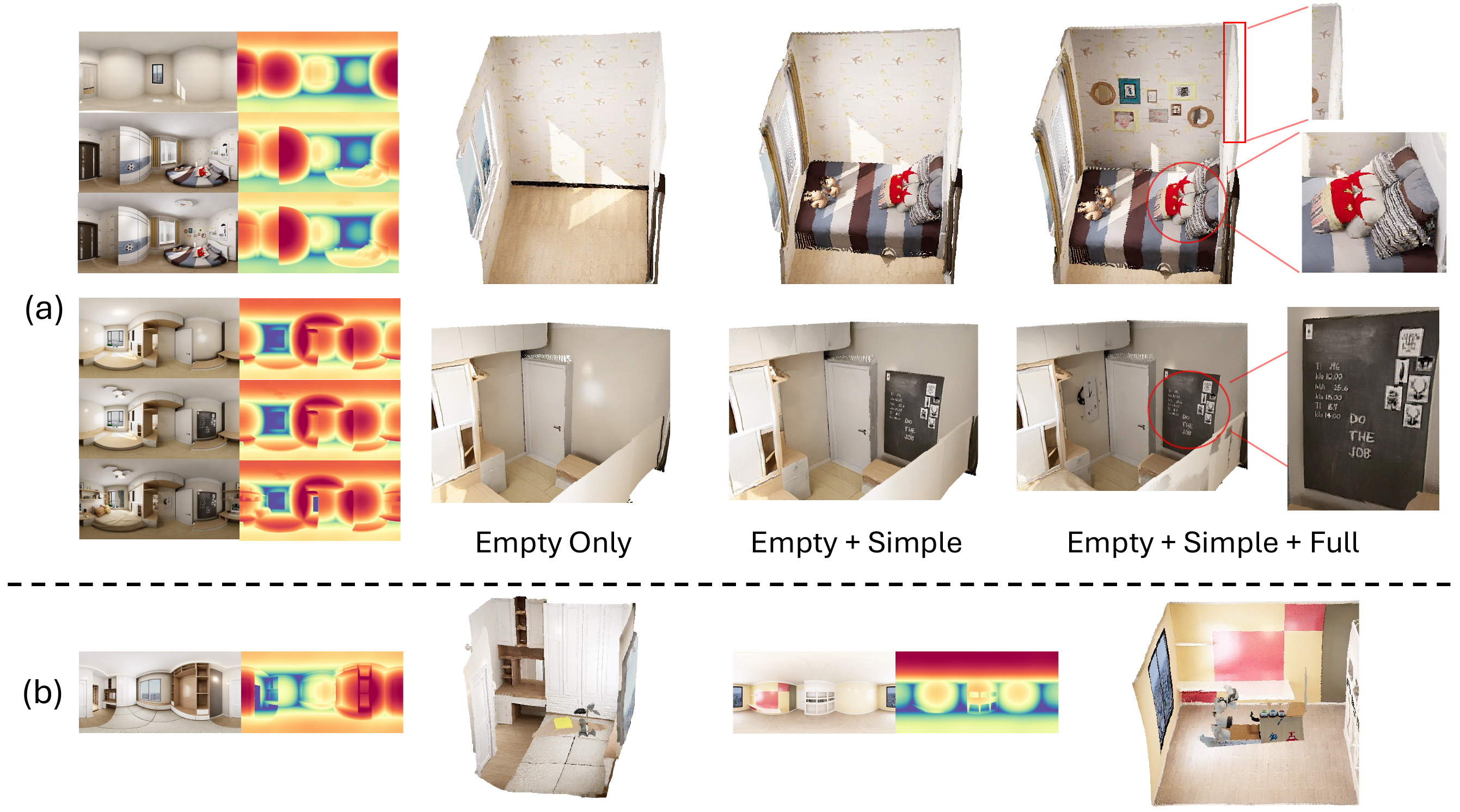}
\caption{More applications of DA$^2$: 
(a) Layered Home Renovation. The three input panoramas correspond to different levels of foreground object complexity, denoted as ``empty'', ``simple'', and ``full''. The zoom-in views show that the reconstructed 3D point clouds from these panoramas remain consistently aligned (primarily in backgrounds). 
(b) Robotics Simulation. The reconstructed 3D point cloud can serve as a practical 3D platform for evaluating robot manipulation performance.}

    \label{fig:application2}
\end{figure}

\subsection{Layered Home Renovation}
As illustrated in Fig.~\ref{fig:application2} (a), given indoor panoramas with three distinct complexity levels---``empty'', ``simple'', and ``full''---the multiple sets of 3D point clouds reconstructed from DA$^2$'s panoramic distance maps exhibit high consistency. They can be seamlessly aligned with fine details.
As demonstrated in the zoom-in regions of Fig.~\ref{fig:application2} (a), the fused point clouds are free of distortions: the text on the blackboard is sharp, and the wall boundaries are consistently aligned.  

\subsection{Robotics Simulation}
Benefiting from DA$^2$'s robust panoramic distance estimation, the reconstructed 3D point cloud can serve as a reliable 3D simulation environment for robot manipulation. As illustrated in Fig.~\ref{fig:application2} (b), it provides a practical 3D platform for simulating and demonstrating robotic tasks.




\vspace{.3cm}

\section{Panoramic Data Curation Engine (More Details)}
\label{supp:sec:data}

As discussed in Sec.~\ref{sec:exp_set}, 6 perspective datasets (Hypersim~\citep{Hypersim},
Virtual-KITTI 2~\citep{vkitti},
MVS-Synth~\cite{MVSSynth},
UnrealStereo4K~\citep{unrealstereo4k},
3D-Ken-Burns~\citep{3dkenburns},
Dynamic Replica~\citep{DynamicReplica}) are transformed into panoramic via the proposed panoramic data curation engine.
The curated datasets are summarized in Tab.~\ref{supp:tab:data}. As shown,
the sampling probabilities are normalized across datasets primarily considering data size to ensure a balanced influence during DA$^2$'s training process.
Each dataset represents a domain, this balanced mixture ensures DA$^2$'s performance will not be over influenced by a few strong datasets, achieving stable scaling behavior across datasets, and optimal cross-domain generalization.
To this end, our data curation engine generates $\sim$543K high-quality panoramic image–depth pairs from perspective data, expanding the total dataset to $\sim$607K samples.
This substantially enriches the quantity and diversity of panoramic data, constructs a solid data foundation for DA$^{2}$, and in turn significantly enhances the zero-shot performance of DA$^{2}$, as demonstrated in Fig.~\ref{fig:scalinglaw} and Tab.~\ref{tab:scalinglaw}.






\begin{table}[!h]
\centering
\footnotesize

\caption{Perspective datasets processed by the panoramic data curation engine. 
For each dataset, the vertical FoV (YFoV) is derived directly from the horizontal FoV (XFoV) as 
$\text{YFoV} = \text{XFoV}\times\frac{H}{W}$, where $(W,H)$ denotes the input panorama's width and height.}

\begin{NiceTabular}{ccccccc}
\toprule
Category& Dataset Name& Abbreviation (Tab.~\ref{tab:scalinglaw}) & Data Size & In-or-outdoor & XFoV& Sam.~Probability\\
\cmidrule(lr){1-7}
Perspective& Hypersim& HPS& 39,649&In & 60$^\circ$& 16.59\%\\
Perspective& Virtual-KITTI 2& VK& 33,580& Out& 80$^\circ$& 14.05\%\\
Perspective& MVS-Synth& MVS& 12,000& Out& 80$^\circ$& 5.02\%\\
Perspective& UnrealStereo4K& US4K& 16,400& Various& 90$^\circ$& 6.86\%\\
Perspective& 3D-Ken-Burns& 3DKB&151,996 & Various& 60$^\circ-$90$^\circ$& 15.91\%\\
Perspective& Dynamic Replica& DR& 289,800& In& 85$^\circ$& 15.16\%\\
\cmidrule(lr){1-7}
Panoramic& Structured3D& S3D& 63,097&In &360$^\circ$ &26.41\% \\
\bottomrule
\end{NiceTabular}
\label{supp:tab:data}
\end{table}


\section{Evaluation Metrics}
\label{supp:sec:metrics}

Concretely, given the predicted panoramic depth $\hat D$ and GT $D^\star$, median alignment is performed on the predicted distance $\hat{D}$ before computing the metrics:
\begin{equation}
\hat{D}^\text{med}=\hat{D}\times\frac{\operatorname{Median}(D^\star)}{\operatorname{Median}(\hat{D})},
\end{equation}
following the evaluation protocols in prior works~\citep{lee2025hush,wang2024depthanywhere,yun2023egformer,yan2025spherefusion,yan2022spheredepth,li2022omnifusion,shen2022panoformer,zhuang2022acdnet,wang2022bifuse++,sun2021hohonet,pintore2021slicenet,jiang2021unifuse,wang2020bifuse,rey2022360monodepth,piccinelli2025unik3d,cao2025panda},
After that, the AbsRel and RMSE are given by:
\begin{equation}
\text{AbsRel}=\frac{1}{|\Omega|}
\sum_{p\in\Omega}\frac{|\hat{D}^\text{med}_p-D^\star_p|}{D^\star_p},\quad\text{RMSE}=\frac{1}{|\Omega|}
\sqrt{\sum_{p\in\Omega}(\hat{D}^\text{med}_p-D^\star_p)^2},
\end{equation}
where $\Omega$ is the set of valid pixels.
$\delta 1$ and $\delta2$ denotes the proportion of pixels satisfying $\text{Max}(D^\star_p/\hat{D}^\text{med}_p, \hat{D}^\text{med}_p/D^\star_p)<1.25$ and $< 1.25^2$ respectively.

\vspace{3mm}
\section{Difference among: Metric \& Scale-invariant (Biased) \& Affine-invariant (Relative)}
\label{supp:depth_category}

\noindent\textbf{Metric and Scale-invariant Depth.}~In depth (or distance) estimation, metric depth $D_\text{metric}$ is the strictest setting, where the predicted values correspond to absolute physical distances and can be directly used to reconstruct a ``real-scale'' point cloud.
Scale-invariant (or biased) depth $D_\text{biased}$ is still strict but slightly more relaxed than metric: predictions include a global bias or shift, but not in the absolute global scale. Although the depths are not metric, the underlying 3D structure is preserved perfectly (Tab.~\ref{supp:tab:concept}), because the global bias or shift is preserved. During training \& evaluation, for scale-invariant depth, median alignment (scale-invariant) is typically adopted to re-scale the underlying 3D structure to real-world size (please see Sec.~\ref{supp:sec:metrics}). For metric depth, no alignment should ideally be required, but median alignment is still commonly applied because absolute scales can be ambiguous
(cameras with different focal lengths can capture visually similar pictures but with substantially different absolute depths)~\citep{hu2024metric3d,yin2023metric3d,piccinelli2025unik3d}.

DA$^2$ focuses on panoramic scale-invariant (or biased) distance estimation for two reasons: 1) like metric distance, scale-invariant distance also preserves the full underlying 3D geometry, and 2) DA$^2$ targets on the strong zero-shot generalization across diverse domains, enforcing absolute scales would introduce significant optimization challenges, as indoor and outdoor scenes differ drastically in scale, making the additional cost outweigh the benefits.

\noindent\textbf{Affine-invariant Depth.}~Affine-invariant (or relative) depth $D_\text{relative}$ is the loosest definition, much more relaxed than either biased or metric depth, preserving only the ``ordering'' of depths (which point is closer or farther). Since neither scale nor shift is preserved, affine-invariant depth $D_\text{relative}$ cannot be used to reconstruct a \textit{reasonable} 3D point cloud (Tab.~\ref{supp:tab:concept}), but it's useful for tasks where only relative geometry matters.
Affine-invariant alignment (scale and shift-invariant) is usually adopted during training \& evaluation of affine-invariant depth estimators.
Concretely, given the predicted $\hat D_\text{relative}$ and GT depth $D^\star$, least squares fitting is performed:
\begin{equation}
\min_{\text{scale, shift}}
\sum_{p\in\Omega}
\bigl\|\,\text{scale}\times(\hat{D}_{\text{relative}, p} + \text{shift})-D^\star_p\,\bigr\|^2_2,\quad
\end{equation}
where $\Omega$ is the set of valid pixels and the aligned predicted depth is: $\hat D^
{\text{aff}}=\text{scale}\times(\hat{D}_{\text{relative}} + \text{shift})$.

The summarized difference is listed in Tab.~\ref{supp:tab:concept}.
Note that for the ``Illustration with $D_\text{metric}$'' of scale-invariant and affine-invariant depth, we only list the most widely adopted formats, passing over other scales for $D_\text{biased}$ and other specific transformations for $D_\text{relative}$ like $\exp(\cdot)$ and $\log(\cdot)$.

\setlength{\tabcolsep}{1.5pt}
\begin{table}[!h]
\centering
\footnotesize
\caption{Summarized difference on depth maps among metric, scale-invariant (biased), and affine-invariant (relative).
Both metric and scale-invariant depth fully preserve the 3D geometry.
Due to the absence of bias or shift, affine-invariant depth is unable to reconstruct an accurate 3D structure.}
\begin{NiceTabular}{cccc}
\toprule
Depth Category& Metric Depth& Scale-invariant Depth & Affine-invariant Depth\\
\cmidrule(lr){1-4}
Illustration with $D_\text{metric}$ & 
$D_\text{metric}$& 
$\frac{D_\text{metric}}{\max(D_\text{metric})}$&
$\frac{D_\text{metric} - \min(D_\text{metric})}{\max(D_\text{metric}) - \min(D_\text{metric})}$\\
\cmidrule(lr){1-4}
\includegraphics[width=0.16\textwidth]{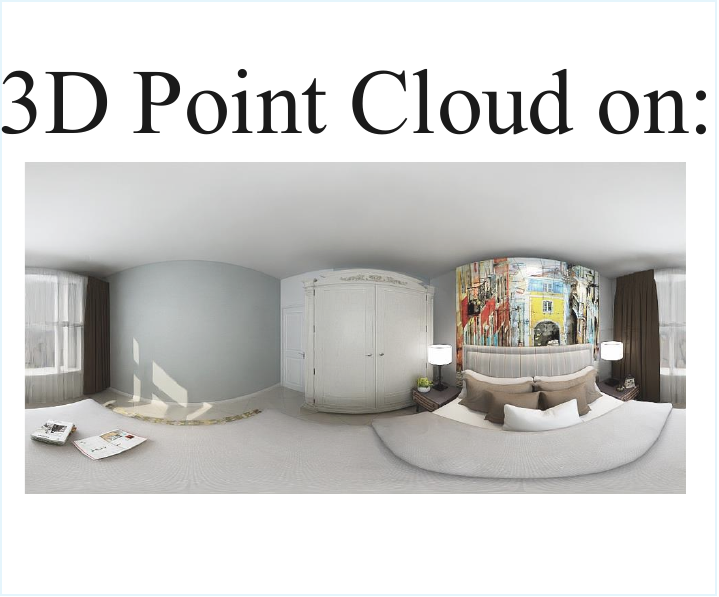}
&\includegraphics[width=0.25\textwidth]{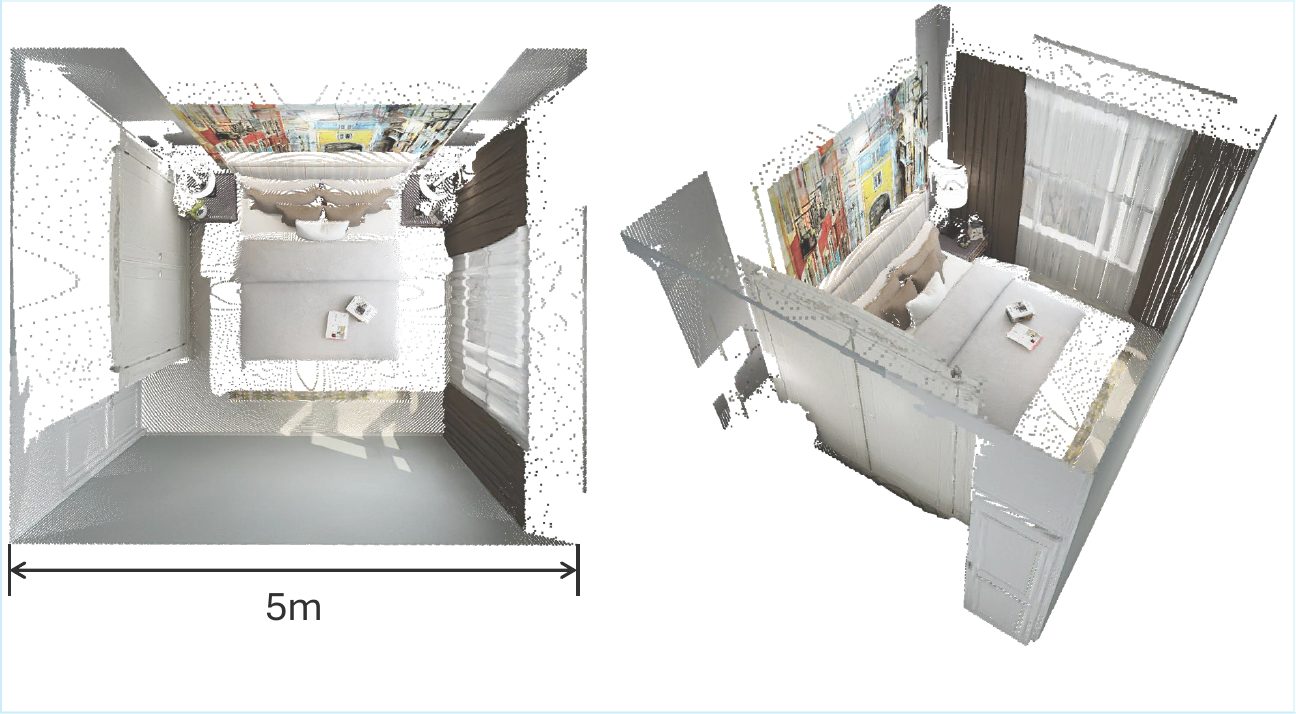} 
& \includegraphics[width=0.25\textwidth]{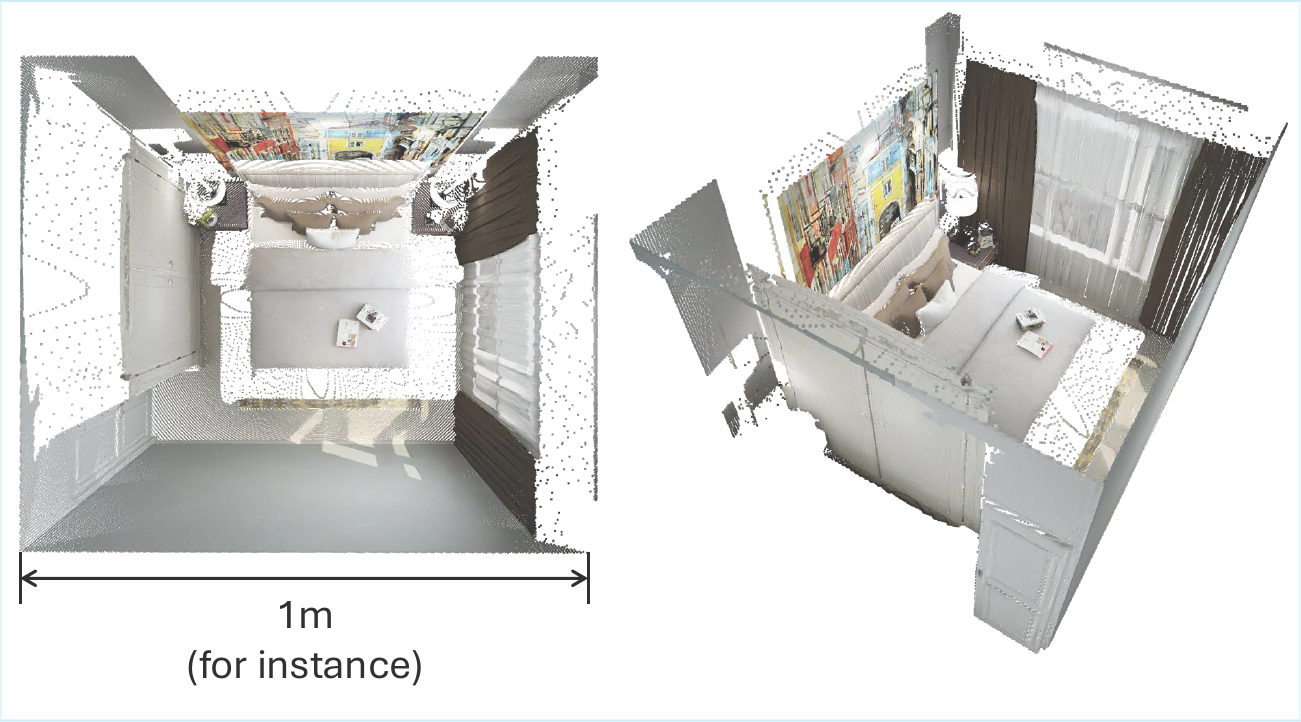} 
& \includegraphics[width=0.25\textwidth]{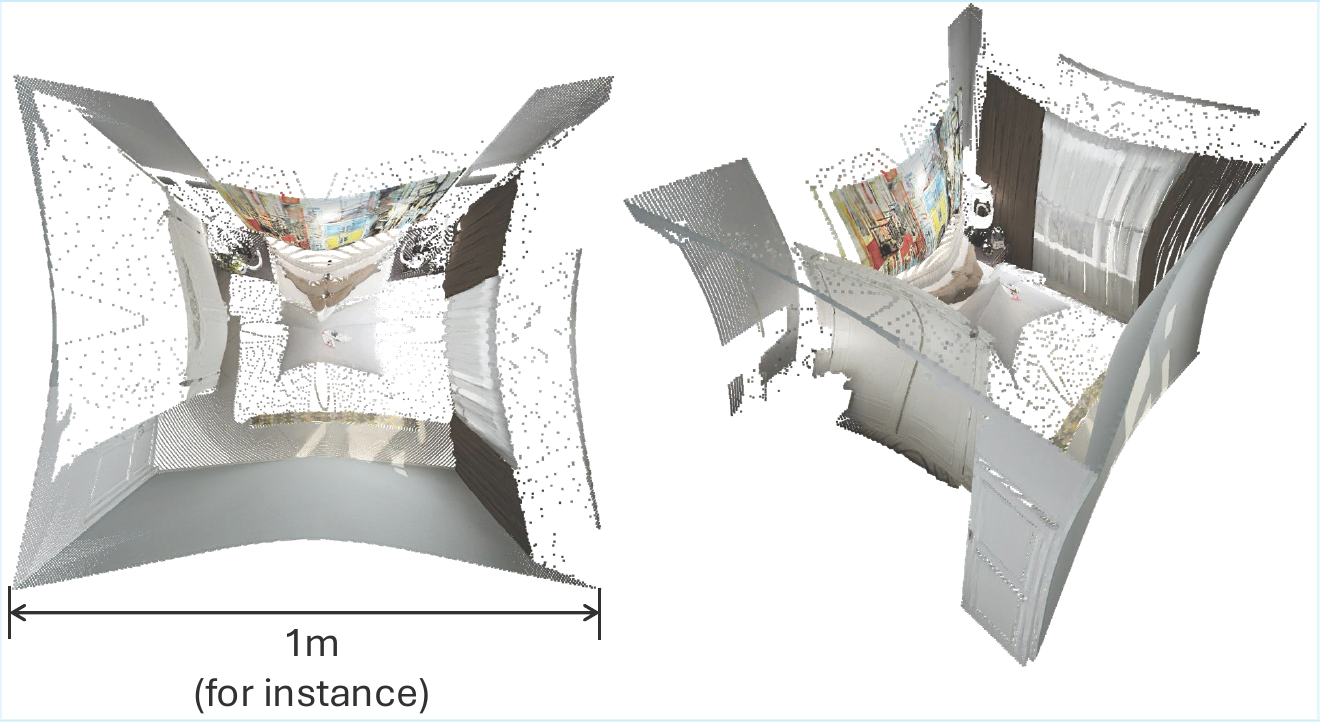} 
\\
\bottomrule
\end{NiceTabular}
\label{supp:tab:concept}
\end{table}



\section{Implementation Details}
\label{supp:sec:impl}

DA$^2$ is implemented in PyTorch~\citep{paszke2019pytorch}.
In SphereViT, the backbone of ``ViT (DINO)'' is initialized from DINOv2-ViT-L~\citep{oquab2023dinov2} with 24 self-attention blocks, following~\citep{he2024lotus,marigold}, to leverage the pre-trained visual priors.
The ``ViT \textbf{w/} $E_\text{sphere}$'' is a lightweight ViT contains only 4 cross-attention blocks.
Training the SphereViT takes $\sim$5,000 optimization iterations on 32 NVIDIA H20 GPUs, with a batch size of 768.
The distributed training is implemented with Accelerate~\citep{accelerate}.
We set $\lambda_\text{dis}=1.0, \lambda_\text{nor}=2.0$ for balanced loss values.
Panoramas and GT depth maps are fed to SphereViT at a resolution of 1024$\times$512.
Please see the sampling probabilities of different data sources in Tab.~\ref{supp:tab:data}.
In the panoramic data curation engine, the FLUX-I2P is fine-tuned on FLUX.1 [dev]~\citep{flux2024}, largely following~\cite{hunyuanworld2025tencent}.
The LoRA rank is set to 256 during the LoRA~\citep{hu2022lora} fine-tuning.
The positive prompt is: \texttt{a clean, realistic, high-quality, high-resolution, panoramic image of a [*] scene}, where \texttt{[*]} is either \texttt{indoor} or \texttt{outdoor}.
The negative prompt is: \texttt{messy, low-quality, blur, noise, low-resolution, abnormal}.
The $\phi_c, \theta_c$ are randomly selected from $\pm$30$^\circ$ and $\pm$15$^\circ$, respectively.
The panoramic out-painting of 543,425 perspective RGB images from various datasets is performed on 64 NVIDIA H20 GPUs and over nearly 9 days.
The running time reported in Fig.~\ref{fig:qc} is tested on a NVIDIA H20 GPU at a resolution of 1024$\times$512, excluding I/O operations.

\section{Prior SoTA Methods for Comparisons}

\noindent\textbf{In-domain Baselines.}~
17 previous in-domain, panoramic depth estimation approaches are selected for the quantitative comparison in Tab.~\ref{tab:main}: HUSH~\citep{lee2025hush}, DepthAnywhere~\citep{wang2024depthanywhere}, Elite360D~\citep{ai2024elite360d}, EGFormer~\citep{yun2023egformer}, SphereFusion~\citep{yan2025spherefusion}, SphereDepth~\citep{yan2022spheredepth}, OmniFusion~\citep{li2022omnifusion}, HRDFuse~\citep{ai2023hrdfuse}, PanoFormer~\citep{shen2022panoformer}, ACDNet~\citep{zhuang2022acdnet}, BiFuse++~\citep{wang2022bifuse++}, HoHoNet~\citep{sun2021hohonet}, SliceNet~\citep{pintore2021slicenet}, UniFuse~\citep{jiang2021unifuse}, BiFuse~\citep{wang2020bifuse}, FCRN~\citep{laina2016deeper}, and OmniDepth~\citep{zioulis2018omnidepth}.

\noindent\textbf{Zero-shot, fusion-based baselines.}~13 zero-shot, fusion-based panoramic depth estimators are selected or implemented.
1 is originally panoramic: 360MonoDepth~\citep{rey2022360monodepth}. The other 16 are prior SoTA perspective depth estimators: Metric3D \& Metric3Dv2~\citep{yin2023metric3d,hu2024metric3d}, VGGT~\citep{wang2025vggt}, MoGe \& MoGev2~\citep{wang2025moge,wang2025moge2}, UniDepth \& UniDepthv2~\citep{piccinelli2024unidepth,piccinelli2025unidepthv2}, ZoeDepth~\citep{bhat2023zoedepth}, DepthAnything \& DepthAnythingv2~\citep{yang2024depthanything,yang2024depthanythingv2}, and Lotus-D \& Lotus-G~\citep{he2024lotus}.
These methods are implemented for panoramic scenarios via multi-view splitting and fusion.

\noindent\textbf{Zero-shot, end-to-end baselines.}~
Prior zero-shot, end-to-end methods are rare, and their performance are limited by the scarcity of high-quality panoramic depth data.
Only 3 methods are compared: UniK3D~\citep{piccinelli2025unik3d}, PanDA~\citep{cao2025panda}, and DepthAnyCamera~\citep{DepthAnyCamera}.
As evident in Tab.~\ref{tab:main}, PanDA predicts affine-invariant (relative) depth, while other methods including DA$^2$ predict at least the scale-invariant (biased) depth.

\bibliography{iclr2026_conference}
\bibliographystyle{iclr2026_conference}

\end{document}